\documentclass{article}

\usepackage[preprint]{neurips_2025}

\usepackage[utf8]{inputenc} 
\usepackage[T1]{fontenc}    
\usepackage{url}            
\usepackage{booktabs}       
\usepackage{amsfonts}       
\usepackage{nicefrac}       
\usepackage{microtype}      

\usepackage{amsmath}
\usepackage{amssymb}
\usepackage{mathtools}
\usepackage{amsthm}
\usepackage{booktabs}
\usepackage{multirow}
\usepackage{adjustbox}
\usepackage{colortbl}

\usepackage{wrapfig}
\usepackage{subcaption}  
\usepackage{xcolor}         
\definecolor{neuripsblue}{HTML}{03A5FC}  
\usepackage[
  colorlinks=true,    
  citecolor=neuripsblue,     
  linkcolor=neuripsblue,    
  urlcolor=blue,      
  pdfborder={0 0 0}   
]{hyperref}
\usepackage{etoc} 
\usepackage{tcolorbox}
\usepackage{booktabs}
\usepackage{xcolor}
\usepackage{colortbl}

\usepackage{fontawesome}  
\usepackage{graphicx}
\usepackage{adjustbox}
\newsavebox{\stripbox}

\definecolor{bestrow}{RGB}{232, 245, 253}
\definecolor{hdrblue}{RGB}{70, 130, 180}
\definecolor{hdrorange}{RGB}{210, 120, 70}
\newcommand{\best}[1]{\textbf{#1}}
\usepackage{graphicx}
\usepackage{pgf}             
\usepackage{enumitem}
\newsavebox{\tabbox}

\definecolor{bestrow}{RGB}{232, 245, 253}
\definecolor{hdrblue}{RGB}{70, 130, 180}
\definecolor{hdrorange}{RGB}{210, 120, 70}
\definecolor{dgood}{RGB}{0,130,60}
\definecolor{dbad}{RGB}{190,30,30}
\newcommand{\dgain}[1]{{\scriptsize\textcolor{dgood}{$+#1$}}}
\newcommand{\dloss}[1]{{\scriptsize\textcolor{dbad}{$-#1$}}}
\newcommand{\dcell}[2]{#1\,{\scriptsize\rlap{#2}\phantom{$+0.000$}}}
\providecommand{\best}[1]{\textbf{#1}}

\title{Tri-Info: Generalizable, Interpretable Failure Prediction for VLA Models via Information Theory}

\author{%
  Jinghan Yang \thanks{HKU Musketeers Foundation Institute of Data Science, email: jinghanyang@connect.hku.hk} \quad
  Yunchao Zhang \quad
  Wang Yuan \quad
  Haolun Wang \\[2pt]
  \bfseries
  Jiaming Zhang \quad
  Zhengyang Hu \quad
  Yanchao Yang \\[6pt]
  InfoBodied AI Lab, The University of Hong Kong \\
}

\begin{document}

\maketitle

\begin{abstract}
Vision-Language-Action (VLA) models are increasingly deployed across diverse tasks, yet they remain black boxes whose physical interactions can cause irreversible harm, making generalizable and interpretable failure detection essential. We observe that successful and failed rollouts carry systematically different information-theoretic signatures. Building on this, we formalize VLA control as a closed-loop information pipeline and derive the Triple Information-theoretic (Tri-Info) signals that capture whether actions remain diverse, temporally consistent, and coupled to state transitions. Across six VLA models and three benchmark environments, Tri-Info matches the strongest baselines in-domain. Moreover, Tri-Info transfers across architectures, environments, and the sim-to-real gap without retraining with labeled data, reaching 70\% accuracy on real-world tasks. This establishes Tri-Info as a simple yet powerful method that not only detects failures with strong cross-domain generalization, but also delivers interpretable diagnostics of the underlying failure modes.
\end{abstract}

\section{Introduction}
Building on large-scale vision-language pretraining~\citep{radford2021learning, radford2019language, brown2020language}, vision-language-action (VLA) models map multimodal inputs directly to robot actions via behavioral cloning~\citep{brohan2022rt, brohan2023rt, kim2024openvla, o2024open}. Yet they fail silently under distributional shift, such as unfamiliar objects~\citep{zhu2025objectvla}, lighting~\citep{jin2025physically, xie2024decomposing}, or tasks~\citep{zhou2025exploring}. Unlike software errors, such failures are often irreversible: a misguided reach can injure a person or shatter a fragile object. Safe deployment therefore requires predicting failures before they unfold~\citep{brunke2022safe, xing2025towards}.
Existing detectors either train classifiers on a policy's internal representations~\citep{Gu2025SAFEMF, xu2025can, romer2025failure}, achieving strong in-domain accuracy, or monitor the temporal consistency of sampled actions~\citep{agia2024unpacking}. The former are tied to one architecture's feature space and must be retrained for each new policy; both compress a rollout into a single score that signals whether it fails but not how. Both also inspect the policy in isolation, whereas failure arises in the closed loop between the policy and its environment.

We study failure from the closed-loop perspective, drawing on information theory: \textbf{effective information transfer is a prerequisite for achieving a goal} ~\citep{shannon1949mathematical}. 
A closed-loop agent, animal or machine, can succeed only if its perception and action remain informationally coupled with its environment: observations must reduce uncertainty about what to do, and actions must in turn shape what is observed next \citep{wiener1948cybernetics, tishby2010information}.
In the perception--action loop $\mathbf{S}_t\to\mathbf{A}_t\to\mathbf{S}_{t+1}\to\mathbf{A}_{t+1}$, the entropy of each variable quantifies how much information it carries, and the mutual information (MI) between variables quantifies how much of it passes along the loop. A successful rollout places three requirements on this flow. First, the agent must retain enough variety in its actions to counter variation in its environment~\citep{ashby1956introduction}. Second, its actions must remain coherent over time, since a physical skill unfolds as a sequence of related motions rather than a series of independent decisions~\citep{zhao2023learning, chi2025diffusion}. Third, its actions must stay grounded in what it observes, since a controller that ignores its observations can no longer correct its course~\citep{touchette2000information}. Failure breaks one of these requirements: action entropy collapses below or surges beyond what the task demands, consecutive actions lose coherence, or actions stop responding to observations. We therefore hypothesize: \emph{Success and failure rollouts differ in how these information-theoretic quantities evolve over time.}

\begin{figure}[t]
    \centering
    \includegraphics[width=0.99\linewidth]{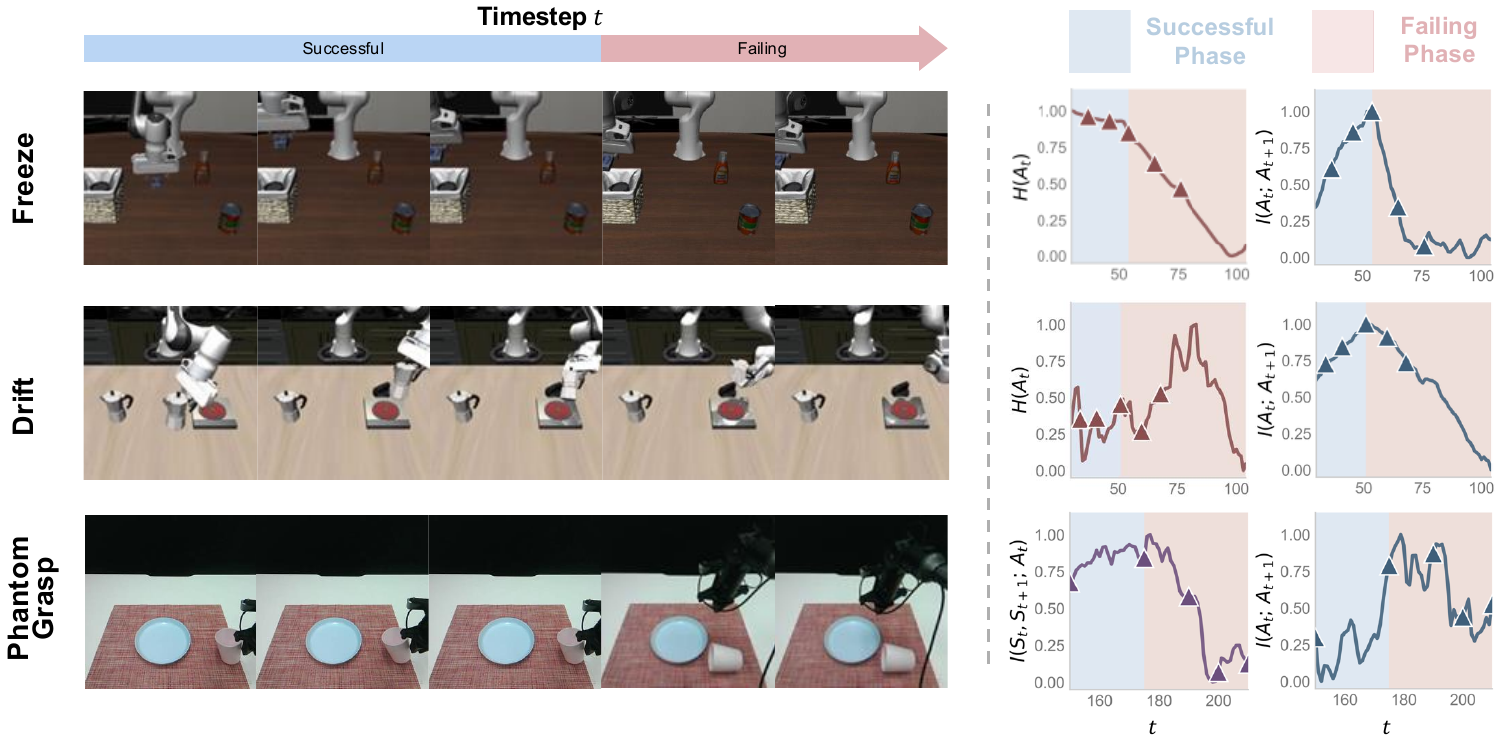}
    \caption{\textbf{Information-theoretic metrics shift at failure onset.}
Representative PI$_0$-LIBERO trajectories under three failure modes: freeze ($\mathbb{H}(\mathbf{A}_t)\!\downarrow$, $\mathbb{I}(\mathbf{A}_t;\mathbf{A}_{t+1})\!\downarrow$), drift ($\mathbb{H}(\mathbf{A}_t)\!\uparrow$, $\mathbb{I}(\mathbf{A}_t;\mathbf{A}_{t+1})\!\downarrow$), and phantom grasp ($\mathbb{I}(\mathbf{S}_t,\mathbf{S}_{t+1};\mathbf{A}_t)\!\downarrow$, $\mathbb{I}(\mathbf{A}_t;\mathbf{A}_{t+1})\!\downarrow$). Left: 5-frame visual strips; right: the corresponding metrics, each shifting sharply as the trajectory crosses from its successful into its failing phase. Different modes leave distinct signatures, motivating the three complementary Tri-Info signals.}    
    \label{fig:figure1}
    \vspace{-3em}
\end{figure}

To test this hypothesis, we formalize VLA-controlled systems as closed-loop information pipelines and derive eight metrics in four diagnostic categories: marginal statistics ($\mathbb{H}(\mathbf{S}_t)$, $\mathbb{H}(\mathbf{A}_t)$), policy coupling ($\mathbb{I}(\mathbf{S}_t; \mathbf{A}_t)$), dynamics ($\mathbb{I}(\mathbf{A}_t; \mathbf{S}_{t+1})$, $\mathbb{I}(\mathbf{A}_t, \mathbf{S}_t; \mathbf{S}_{t+1})$, $\mathbb{I}(\mathbf{S}_t, \mathbf{S}_{t+1}; \mathbf{A}_t)$), and temporal coherence ($\mathbb{I}(\mathbf{S}_t; \mathbf{S}_{t+1})$, $\mathbb{I}(\mathbf{A}_t; \mathbf{A}_{t+1})$), each estimated over a sliding window of the rollout.
Removing redundancy and favoring action-centric metrics yields three complementary signals, the \textbf{Tri}ple \textbf{Info}rmation-theoretic (\textbf{Tri-Info}) metrics $\{\mathbb{H}(\mathbf{A}_t),\, \mathbb{I}(\mathbf{A}_t;\mathbf{A}_{t+1}),\, \mathbb{I}(\mathbf{S}_t,\mathbf{S}_{t+1};\mathbf{A}_t)\}$.
They describe action from three angles: how spread the policy's actions are, how much of that uncertainty is explained by its own recent actions, and how much by what it observes. \textbf{A healthy policy thus acts with moderate entropy that is grounded in both its history and its observations.} 
On top of them, we build a Tri-Info detector that fuses one GRU per signal to model how these signals evolve over a rollout. 

\begin{figure}[t]
    \centering
    \includegraphics[width=0.99\linewidth]{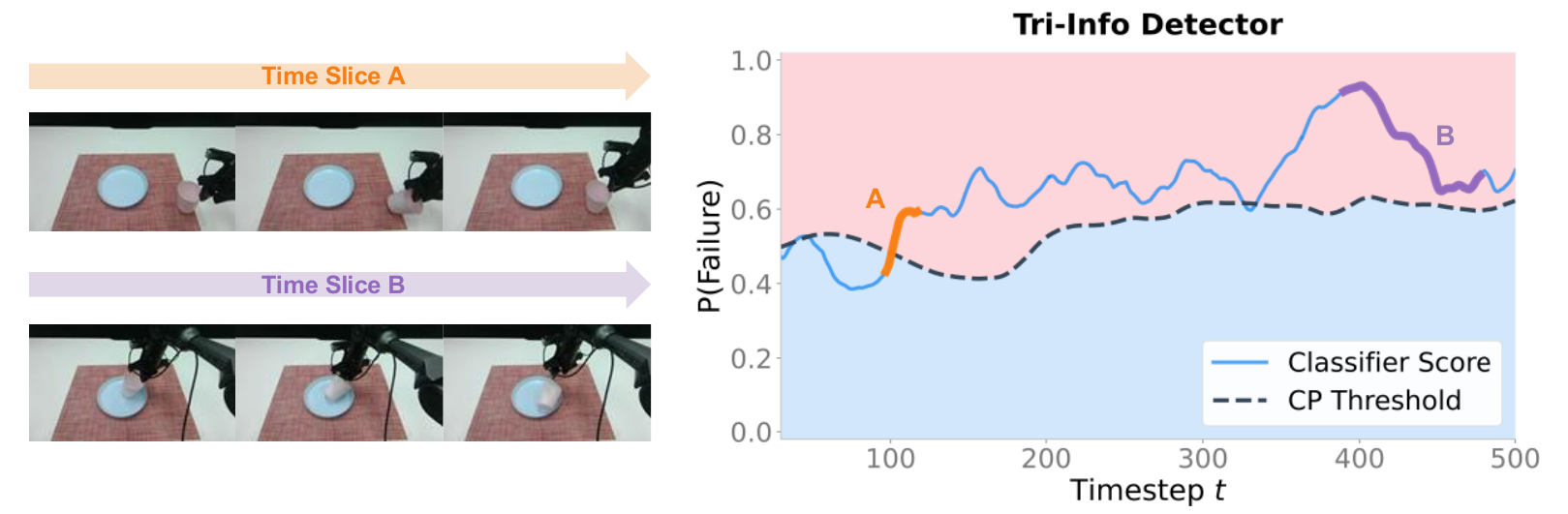}
    \caption{\textbf{The Tri-Info detector raises $P(\mathrm{Failure})$ ahead of each dangerous event on ACT-ALOHA-real.}
The fused score (blue) crosses the threshold (dashed) on two near-tipping events ($t{=}97$–$117$, $390$–$477$, marked by image insets).}
    \label{fig:failure_trajectory_demo}
    \vspace{-1em}
\end{figure}

Our experiments support the hypothesis and reveal two further properties. First, the metrics are \emph{interpretable}: different failure modes shift different metrics (Figure~\ref{fig:figure1}). In a \emph{freeze}, the robot stalls and $\mathbb{H}(\mathbf{A}_t)$ collapses. In a \emph{drift}, the robot abandons its goal and moves erratically, and $\mathbb{H}(\mathbf{A}_t)$ surges instead. In a \emph{phantom grasp}, the robot acts as if holding an object it never grasped, and $\mathbb{I}(\mathbf{S}_t,\mathbf{S}_{t+1};\mathbf{A}_t)$ drops. The pattern of shifts thus indicates not only that a rollout is failing but also how, and the detector's alarms align with visually identifiable dangerous events (Figure~\ref{fig:failure_trajectory_demo}). Second, the metrics are \emph{generalizable}: Tri-Info matches the strongest baselines in domain and retains its accuracy across VLA architectures, task environments, and the sim-to-real gap, without labeled failure data from the target domain. 
We attribute this to the metrics summarizing statistical dependencies among states and actions.
We evaluate Tri-Info on various VLA models, simulated benchmarks, and real-world robot tasks, and the summarized contributions are:

\noindent\textbf{1. Information-theoretic framework.} We formalize VLA control as a closed-loop information pipeline and propose three complementary Tri-Info signals for failure prediction.

\noindent\textbf{2. Interpretable failure signatures.} We show that freeze, drift, and phantom grasp leave corresponding signatures on the Tri-Info signals.

\noindent\textbf{3. Cross-domain generalization.} We show that Tri-Info transfers across architectures, environments, and the sim-to-real gap without labeled failure data from the target domain.

\section{Related Work}
\noindent\textbf{Information-Theoretic Analysis in Robotic Control.}
Information theory has long informed robotic control. Empowerment, the channel capacity between actions and future states~\citep{klyubin2005empowerment}, quantifies an agent's ability to influence its environment; variational estimators made it tractable~\citep{mohamed2015variational}, enabling applications in exploration~\citep{houthooft2016vime} and unsupervised skill discovery~\citep{eysenbach2018diversity, sharma2019dynamics}. The information bottleneck~\citep{tishby2000information} has been used to identify decision states for structured exploration and transfer~\citep{goyal2019infobot}.

These ideas have recently entered robotic learning. \citet{hejna2025robot} filters low-quality training trajectories using state--action mutual information, \citet{bai2025rethinking} applies the information bottleneck to behavioral cloning, and the multimodal information bottleneck~\citep{you2024multimodal} preserves task-relevant information across sensor modalities. \citet{Li2025InformationTheoreticGF} uses entropy and mutual information to identify kinematically active scene elements for dual-arm policy transfer, and contact-aware Fisher information maximization~\citep{sathyanarayan2025behavior} synthesizes information-rich contact behaviors. These works apply information measures to training and representation learning; in contrast, we provide the first taxonomy of information-theoretic metrics for VLA failure prediction at deployment, organized around the perception--action loop and spanning marginal statistics, policy coupling, dynamics, and temporal coherence. This grounds detection in interpretable signals that remain comparable across architectures and yields mechanistic insight unavailable through black-box methods.

\vspace{0.1em}
\noindent\textbf{Predictive Failure Detection for Robotic Systems.}
Work on safe reinforcement learning targets training-time safety through constrained optimization~\citep{achiam2017constrained}, human intervention~\citep{saunders2018trial}, or learned recovery~\citep{thananjeyan2021recovery}. For VLA models trained by behavioral cloning, the challenge shifts to deployment-time monitoring~\citep{brunke2022safe, gu2024safe_survey}, where recent detectors divide by classifier input: \emph{embedding-based} methods train on internal VLA representations, while \emph{score-based} methods compute a scalar signal and detect on top of it.
Among embedding-based methods, SAFE~\citep{Gu2025SAFEMF} trains classifiers on VLA latents with functional conformal calibration, generalizing across tasks but not architectures; Fail-Detect~\citep{xu2025can} casts detection as sequential out-of-distribution (OOD) detection without requiring failure data; and FIPER~\citep{romer2025failure} fuses perception-level OOD via random network distillation with action-chunk entropy. All three attain strong in-domain accuracy but are tied to a specific architecture and cannot transfer without retraining. The score-based STAC~\citep{agia2024unpacking} instead measures temporal consistency via the maximum mean discrepancy between overlapping segments of sampled action chunks. It is the closest in spirit to our $\mathbb{I}(\mathbf{A}_t;\mathbf{A}_{t+1})$, but a distance captures consistency alone, whereas $\mathbb{I}(\mathbf{A}_t;\mathbf{A}_{t+1})=\mathbb{H}(\mathbf{A}_{t+1})-\mathbb{H}(\mathbf{A}_{t+1}\mid\mathbf{A}_t)$ captures both action diversity and consistency on executed actions and remains compatible with deterministic VLA inference. 
Our metrics combine the cross-architecture comparability that embedding-based methods lack, with the diagnostic decomposition that a single consistency score cannot
provide.

\section{Method}
\label{sec:method}
\vspace{-0.5em}
\subsection{Problem Formulation}
\label{sec:problem}

\begin{wrapfigure}{r}{0.45\textwidth}
\vspace{-7mm}
\centering
\includegraphics[width=\linewidth]{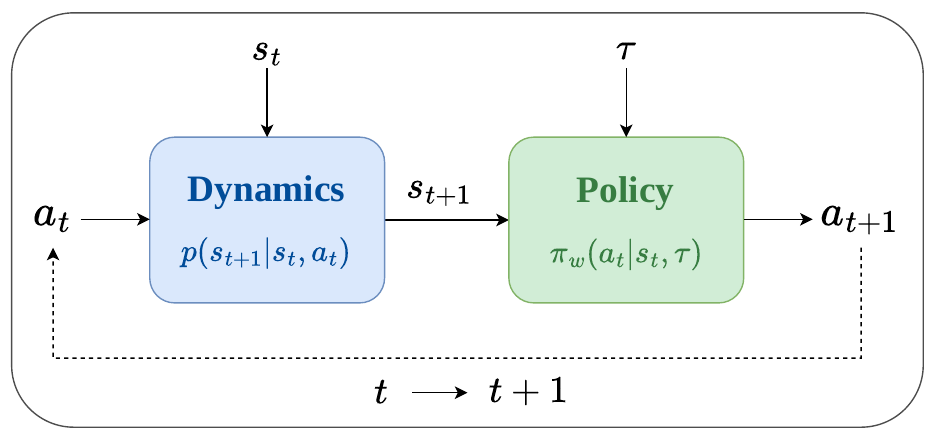}
\vspace{-5mm}
\caption{VLA control as a closed-loop information processing pipeline over $(\mathbf{S}_t, \mathbf{A}_t, \mathbf{S}_{t+1}, \mathbf{A}_{t+1})$ -- the scaffold for the eight derived element metrics.}
\label{fig:flowchar_robot_sys}
\vspace{-0.7em}
\end{wrapfigure}
We model a VLA-controlled system as a tuple $(\mathcal{S}, \mathcal{A}, \mathcal{T}, p, \pi_w)$: $\mathcal{S}$ is the observation space, where $s_t$ comprises the camera images and the proprioception $q_t$, $\mathcal{A}$ the space of low-level controls $a_t$, and $\mathcal{T}$ the space of language instructions $\tau$. At each step $t$, the policy $\pi_w(a_t \mid s_t, \tau)$ emits an action and the environment transitions via $p(s_{t+1}\mid s_t, a_t)$. From $(s_0,\tau)$, interaction induces a trajectory $\xi=\{(s_1,a_1),\ldots,(s_T,a_T)\}$ with a failure label $f^\xi\in\{0,1\}$ ($f^\xi{=}1$ denotes failure).

We extract a state embedding $\mathbf{s}_t=\phi_w(s_t)$ from the state encoder and an action embedding $\mathbf{a}_t=\psi_w(s_t,\tau)$ from the layer preceding the action decoder; for flow-matching models, which lack a discrete decoder, we use the velocity field at the final integration step. Throughout, $\mathbf{S}_t,\mathbf{A}_t,\mathbf{S}_{t+1},\mathbf{A}_{t+1}$ denote the random variables of the current- and next-step embeddings.

\vspace{0.1em}
\noindent\textbf{Online detection objective.}
At each step $t$ the detector reads a scalar statistic $x_t$ and outputs a failure probability $P^\xi_t(\mathrm{Failure}\mid x_t)\in[0,1]$. We seek a detector whose score stays below a time-varying threshold $\theta(t)$ throughout successful trajectories yet exceeds $\theta(t)$ as early as possible on failed ones, enabling timely intervention.

\subsection{Information dynamics in the perception--action loop}
\label{sec:metrics}
Whether a robot completes its task depends on how it processes information
at every step. 
We characterize this with an information profile: the entropy of each
variable in the perception--action loop over
$(\mathbf{S}_t,\mathbf{A}_t,\mathbf{S}_{t+1},\mathbf{A}_{t+1})$ and the
mutual information among them, i.e., how uncertain each variable is and how
strongly the variables depend on one another
(Figure~\ref{fig:flowchar_robot_sys}).
In a successful rollout, actions vary
to a normal degree, stay consistent across steps, and respond to
observations. Failure breaks one of these properties: action entropy
collapses or surges, consecutive actions lose
consistency, or actions stop responding to observations. 
Our central hypothesis is therefore
that \emph{success and failure rollouts differ in how these information-theoretic quantities evolve over time.} 
We thus estimate the information
over a sliding time-step window and track how it changes during execution.

\noindent\textbf{Information Metrics over Sliding Window.}
At each step $t\ge W$, we collect the $W$ most recent executed transitions,
\begin{equation}
\mathbf{W}_t=\{(\mathbf{s}_i,\mathbf{a}_i,\mathbf{s}_{i+1},\mathbf{a}_{i+1})\}_{i=t-W}^{t-1},
\end{equation}
and treat them as samples from the joint distribution of
$(\mathbf{S}_t,\mathbf{A}_t,\mathbf{S}_{t+1},\mathbf{A}_{t+1})$ at step $t$. Each entropy or
mutual-information quantity is estimated from these samples; as the window
slides, it yields an information-theoretic sequence $x_t^{(m)}$ that traces the information
change over the rollout. The window length $W$ balances estimation variance
against responsiveness to change (Appendix \ref{app:mi_estimation}). We derive entropy ($\mathbb{H}$) and mutual-information ($\mathbb{I}$)
quantities over $\mathbf{S}_t$, $\mathbf{A}_t$, $\mathbf{S}_{t+1}$, and
$\mathbf{A}_{t+1}$, yielding eight metrics, $m=1,\dots,8$, in four
diagnostic categories:
\begin{itemize}[leftmargin=1.2em,itemsep=1pt,topsep=2pt,parsep=0pt]
\item \textit{Marginal statistics.} $\mathbb{H}(\mathbf{S}_t)$ and
$\mathbb{H}(\mathbf{A}_t)$ measure the diversity of visited states and
executed actions within the window.
\item \textit{Policy coupling.} $\mathbb{I}(\mathbf{S}_t;\mathbf{A}_t)$
\citep{hejna2025robot} measures how strongly actions depend on the current
observation, diagnosing observation-independent behavior.
\item \textit{Dynamics.} $\mathbb{I}(\mathbf{A}_t;\mathbf{S}_{t+1})$, the
quantity whose maximum over action distributions defines empowerment
\citep{klyubin2005empowerment}, measures the policy's influence on the next
state; $\mathbb{I}(\mathbf{S}_t,\mathbf{A}_t;\mathbf{S}_{t+1})$ measures
forward predictability; and $\mathbb{I}(\mathbf{S}_t,\mathbf{S}_{t+1};\mathbf{A}_t)$
\citep{kim2018emi} measures how well actions can be inferred from the state
transitions they produce. These three metrics with $\mathbb{I}(\mathbf{S}_t;\mathbf{A}_t)$ form state-action coupling metrics.
\item \textit{Temporal coherence.} $\mathbb{I}(\mathbf{S}_t;\mathbf{S}_{t+1})$
\citep{zhou2024maxmi} measures dependence between consecutive states, and
$\mathbb{I}(\mathbf{A}_t;\mathbf{A}_{t+1})$ measures dependence between
consecutive actions.
\end{itemize}

\begin{wrapfigure}{r}{0.45\textwidth}
\vspace{-0.1em}
\centering
\includegraphics[width=0.99\linewidth]{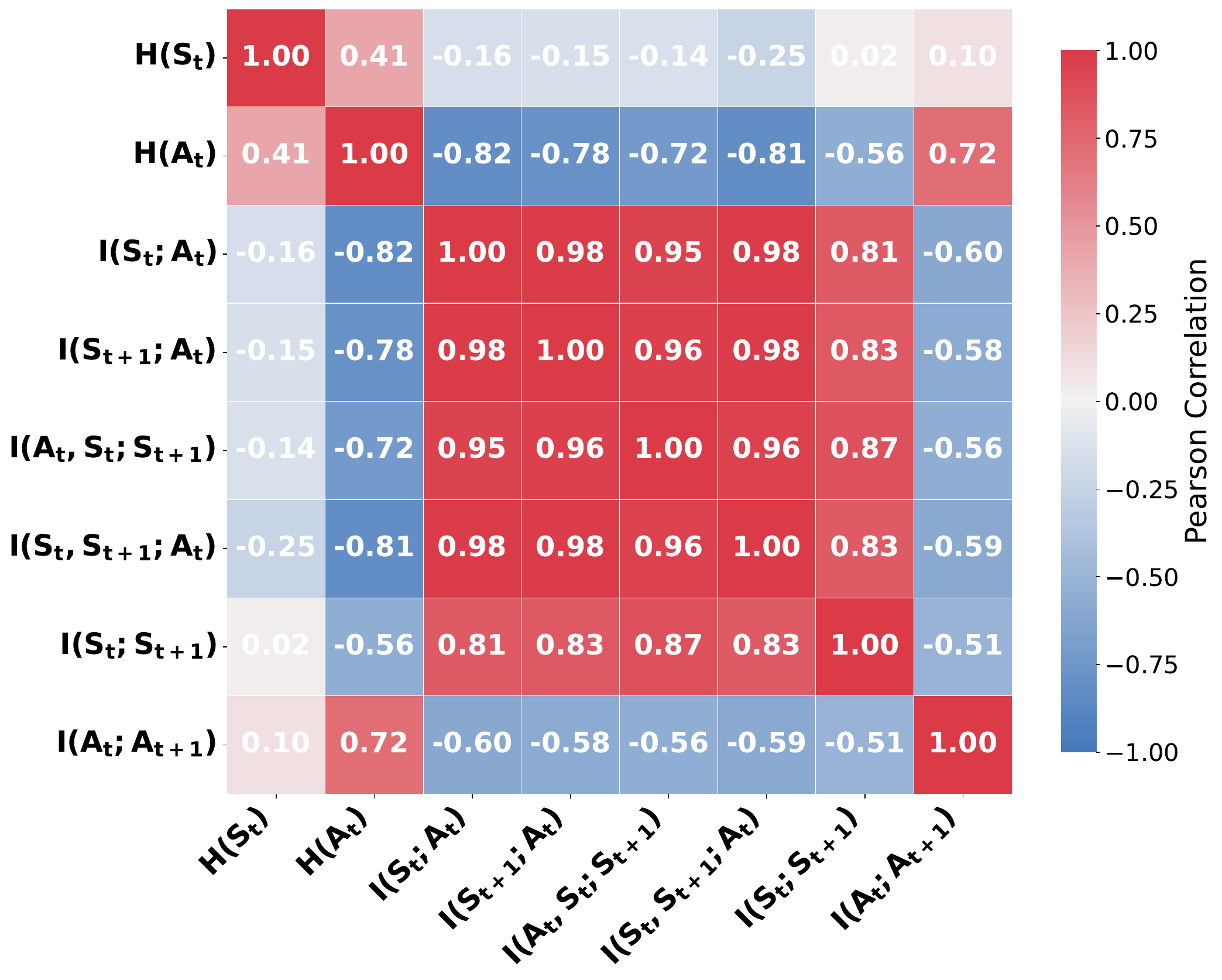}
\caption{\textbf{Pearson correlation of the eight metrics}, pooled over all model-environment combinations. 
The state-action coupling cluster is highly redundant ($r>=0.95$).}
\label{fig:mi_correlation}
\vspace{-1.3em}
\end{wrapfigure}

\noindent\textbf{From eight candidates to three complementary signals.}
The eight metrics are highly redundant, so we seek the smallest subset that still spans the failure modes, guided by the pooled correlation analysis (Figure~\ref{fig:mi_correlation}). \emph{(i)~Eliminate redundancy:} the four state--action coupling metrics are near-perfectly correlated ($r\geq0.95$), so a single representative $\mathbb{I}(\mathbf{S}_t,\mathbf{S}_{t+1};\mathbf{A}_t)$ suffices for the whole cluster. \emph{(ii)~Prefer action-centric metrics}, since actions reflect policy behavior more directly than states; we therefore add $\mathbb{H}(\mathbf{A}_t)$ and $\mathbb{I}(\mathbf{A}_t;\mathbf{A}_{t+1})$, which lie outside this cluster, giving the \textbf{Tri-Info} metrics:
\begin{equation}
\big\{\,\mathbb{H}(\mathbf{A}_t),\;\; \mathbb{I}(\mathbf{A}_t;\mathbf{A}_{t+1}),\;\; \mathbb{I}(\mathbf{S}_t,\mathbf{S}_{t+1};\mathbf{A}_t)\,\big\}.
\end{equation}
These three signals are complementary because they describe action
from three perspectives. 
$\mathbb{H}(\mathbf{A}_t)$ measures how widely the executed actions spread within the window; $\mathbb{I}(\mathbf{A}_t; \mathbf{A}_{t+1})$ measures how predictable each action is from the previous one; and $\mathbb{I}(\mathbf{S}_t, \mathbf{S}_{t+1}; \mathbf{A}_t)$ measures how much of the action is explained by the observed state transition.
A healthy policy acts with moderate
entropy that is grounded in both its history and its observations. Each failure
mode breaks this in a different way. In a \emph{freeze}, action entropy
collapses. In a \emph{drift}, entropy surges while the part explained by
history falls, so the actions become erratic. 
In a \emph{phantom grasp}, the robot moves continuously, but actions are no longer driven by the observations
(Figure~\ref{fig:figure1}).
An exhaustive search over all $2^8-1$ non-empty subsets independently recovers
this triplet as the top set of size 3 by area under the ROC curve (AUC), with no
gain beyond (Appendix~\ref{app:multi-feature}).

\noindent\textbf{Estimation and normalization.}
For the high-dimensional continuous embeddings we use the $k$-NN MI estimator~\citep{hejna2025robot, kraskov2004estimating}
and the Kozachenko--Leonenko entropy estimator~\citep{kozachenko1987sample}.
Because $k$-NN estimators degrade as dimensionality grows, we do not estimate on
the raw embeddings: we first compress $\mathbf{s}_t$ and $\mathbf{a}_t$ to a
matched latent with a variational autoencoder (VAE)~\citep{hejna2025robot}, and estimate all quantities on its posterior
means (details are in Appendix~\ref{app:mi_estimation}). Since the read-out dimensionality is
fixed by the architecture, this projection is fitted per model--dataset domain;
its objective is purely a reconstruction term plus a KL penalty, so it never
sees a failure label. 
Each metric is finally z-normalized as
$z_t=(x_t-\mu)/\sigma$ with statistics computed on the source training split
alone.

\subsection{Online Detection Framework}
\label{sec:detection_framework}
We turn the metrics into a real-time detector with two components: a per-metric temporal model and a time-varying threshold.

\noindent\textbf{Per-metric GRU detector.}
Failure signatures span many consecutive timesteps (Figure~\ref{fig:figure1}), so for each metric we train a single-input GRU over its normalized sequence $\{z_t\}$,
\begin{equation}
\mathbf{h}_t = \mathrm{GRU}(z_t, \mathbf{h}_{t-1}), \qquad
P^\xi_t(\mathrm{Failure}\mid x_t) = \sigma\!\left(\mathrm{MLP}(\mathbf{h}_t)\right) \in [0,1].
\end{equation}
We supervise every timestep with the trajectory-level label $f^\xi$ via BCE,
\begin{equation}
\mathcal{L} = \tfrac{1}{T}\sum_{t=1}^{T}\operatorname{BCE}\!\left(P^\xi_t(\mathrm{Failure}\mid x_t),\,f^\xi\right),
\end{equation}
rather than per-segment, because precise failure onset is semantically ambiguous -- a grasp may succeed or fail at sub-millimeter scales under a visually identical approach. Tying each timestep to the outcome pushes the score toward $1$ as early as possible on failures; empirically, scores stay low at onset and rise only as signatures emerge (Figure~\ref{fig:suc_fail_score}), confirming that ambiguous early timesteps do not corrupt learning.

\vspace{0.1em}
\noindent\textbf{Late mean-probability fusion.}
The three signals target different failure modes (Figure~\ref{fig:figure1}), so we train one GRU per signal \emph{independently} and aggregate only at the output:
\begin{equation}
\bar{P}^\xi_t \;=\; \tfrac{1}{3}\sum_{m\in\mathcal{M}} P^\xi_t\!\left(\mathrm{Failure}\mid x^{(m)}_t\right),
\qquad \mathcal{M}=\{\mathbb{H}(\mathbf{A}_t),\,\mathbb{I}(\mathbf{A}_t;\mathbf{A}_{t+1}),\,\,\mathbb{I}(\mathbf{S}_t,\mathbf{S}_{t+1};\mathbf{A}_t)\}.
\end{equation}
Because the detectors share no parameters or gradients, each tunes its own hyperparameters and signals can be added or dropped at fusion time. The fused score is z-normalized again with training-set statistics before calibration.


\vspace{0.5em}
\noindent\textbf{Threshold calibration.}
Failure is declared at the first timestep where the score exceeds a calibrated threshold $\theta(t)$: $t_{\text{detect}} = \min \{ t \mid \hat{f}^\xi_t > \theta(t) \}$. 
A fixed threshold is inadequate because the success-score distribution shifts over execution.
We adopt Functional Conformal Prediction \citep{diquigiovanni2021importance, Gu2025SAFEMF}, which constructs time-varying thresholds by modeling the temporal distribution of scores from successful calibration trajectories. 
Score sequences from successful trajectories are interpolated to a common normalized time grid $t \in [0,1]$, and the threshold at each point is computed as the corrected $(1-\alpha)$ quantile:
\begin{equation}
\theta(t) = \text{Quantile}_{1-\alpha}\left({\bar{P}^\xi_t}: {\xi \in \mathcal{D}_{\text{cal}}^{+}}\right)
\end{equation}
where $\mathcal{D}_{\text{cal}}^{+}$ denotes the set of successful calibration trajectories. Intuitively, $\theta(t)$ answers how high can the score get at this stage in a successful trajectory?
If a new trajectory's score exceeds $\theta(t)$, it falls outside the $(1-\alpha)$ statistical envelope at the current point for successful behavior and is flagged as a failure. 
We characterize the impact of $\alpha$ empirically in Appendix~\ref{app:Experiment-Details}.

\begin{table*}[!t]
\centering
\newcommand{\bS}{\mathbf{S}}
\newcommand{\bA}{\mathbf{A}}
\newcommand{\bI}{\mathbb{I}}
\newcommand{\bH}{\mathbb{H}}
\newcommand{\ms}[2]{#1$_{\pm #2}$}
\newcommand{\msb}[2]{\textbf{#1}$_{\pm #2}$}
\newcolumntype{B}{>{\columncolor{blue!8}}c}
\caption{Single-metric failure prediction mean balanced accuracy (mean$_{\pm\text{std}}$ over three seeds). Every metric exceeds chance with per-timestep logistic regression (avg.\ 0.65--0.75); temporal modeling via GRU lifts all to 0.72--0.77. Shaded metrics are the components of Tri-Info. Best value per row in \textbf{bold}.}
\label{tab:single_feature}
\renewcommand{\arraystretch}{1.15}
\setlength{\tabcolsep}{5pt}
\resizebox{0.99\textwidth}{!}{%
\begin{tabular}{@{}ll c ccB cB cB@{}}
\toprule
& & \textbf{Policy} & \multicolumn{3}{c}{\textbf{Dynamics}} & \multicolumn{2}{c}{\textbf{Temporal}} & \multicolumn{2}{c}{\textbf{Entropy}} \\
\cmidrule(lr){3-3} \cmidrule(lr){4-6} \cmidrule(lr){7-8} \cmidrule(l){9-10}
\textbf{Environment} & \textbf{Model} & $\mathbb{I}(\bS_t;\bA_t)$ & $\mathbb{I}(\bS_{t+1};\bA_t)$ & $\mathbb{I}(\bA_t,\bS_t;\bS_{t+1})$ & $\mathbb{I}(\bS_t,\bS_{t+1};\bA_t)$ & $\mathbb{I}(\bS_t;\bS_{t+1})$ & $\mathbb{I}(\bA_t;\bA_{t+1})$ & $\mathbb{H}(\bS_t)$ & $\mathbb{H}(\bA_t)$ \\
\midrule
\multicolumn{10}{c}{\textit{Logistic Regression}} \\
\midrule
\multirow{2}{*}{LIBERO-10} & PI$_{0}$ & \ms{0.53}{0.01} & \ms{0.55}{0.03} & \ms{0.56}{0.03} & \ms{0.55}{0.03} & \ms{0.71}{0.02} & \ms{0.82}{0.00} & \ms{0.83}{0.01} & \msb{0.86}{0.02} \\
 & PI$_{0.5}$ & \ms{0.83}{0.03} & \ms{0.81}{0.03} & \ms{0.81}{0.03} & \ms{0.84}{0.03} & \ms{0.78}{0.02} & \ms{0.79}{0.00} & \ms{0.81}{0.03} & \msb{0.88}{0.01} \\
\addlinespace[2pt]
\multirow{3}{*}{CALVIN} & Flower & \ms{0.55}{0.02} & \ms{0.57}{0.01} & \ms{0.90}{0.02} & \msb{0.93}{0.02} & \ms{0.53}{0.03} & \ms{0.50}{0.01} & \ms{0.61}{0.04} & \ms{0.55}{0.03} \\
 & UniVLA & \ms{0.81}{0.03} & \ms{0.67}{0.03} & \ms{0.64}{0.01} & \ms{0.57}{0.02} & \ms{0.73}{0.03} & \ms{0.54}{0.01} & \ms{0.68}{0.07} & \msb{0.92}{0.01} \\
 & GR-1 & \ms{0.81}{0.10} & \ms{0.81}{0.13} & \ms{0.91}{0.02} & \msb{0.97}{0.01} & \ms{0.68}{0.02} & \ms{0.74}{0.02} & \ms{0.76}{0.05} & \ms{0.55}{0.05} \\
\addlinespace[2pt]
ALOHA-transfer & ACT & \ms{0.59}{0.06} & \ms{0.59}{0.06} & \ms{0.50}{0.00} & \ms{0.58}{0.11} & \msb{0.91}{0.03} & \msb{0.91}{0.05} & \ms{0.68}{0.09} & \ms{0.61}{0.03} \\
\addlinespace[2pt]
ALOHA-insertion & ACT & \ms{0.60}{0.01} & \ms{0.61}{0.09} & \ms{0.57}{0.05} & \ms{0.57}{0.05} & \ms{0.66}{0.01} & \ms{0.57}{0.02} & \ms{0.58}{0.01} & \msb{0.80}{0.07} \\
\addlinespace[2pt]
ALOHA-real-cup & ACT & \ms{0.64}{0.02} & \ms{0.64}{0.03} & \ms{0.81}{0.01} & \ms{0.66}{0.01} & \ms{0.69}{0.04} & \ms{0.70}{0.01} & \msb{0.83}{0.02} & \ms{0.80}{0.04} \\
\addlinespace[2pt]
ALOHA-real-drawer & PI$_{0.5}$ & \ms{0.63}{0.08} & \ms{0.60}{0.10} & \ms{0.56}{0.13} & \ms{0.60}{0.13} & \ms{0.64}{0.01} & \ms{0.63}{0.04} & \ms{0.71}{0.00} & \msb{0.76}{0.02} \\
\multicolumn{2}{l}{\textbf{Average}} & 0.67 & 0.65 & 0.70 & 0.70 & 0.70 & 0.69 & 0.72 & \textbf{0.75} \\
\midrule
\multicolumn{10}{c}{\textit{GRU}} \\
\midrule
\multirow{2}{*}{LIBERO-10} & PI$_{0}$ & \ms{0.70}{0.03} & \ms{0.71}{0.02} & \ms{0.72}{0.04} & \ms{0.73}{0.01} & \ms{0.84}{0.01} & \ms{0.84}{0.02} & \ms{0.83}{0.02} & \msb{0.87}{0.02} \\
 & PI$_{0.5}$ & \ms{0.89}{0.02} & \ms{0.86}{0.02} & \ms{0.87}{0.01} & \msb{0.90}{0.01} & \ms{0.85}{0.03} & \ms{0.86}{0.01} & \ms{0.86}{0.02} & \ms{0.87}{0.05} \\
\addlinespace[2pt]
\multirow{3}{*}{CALVIN} & Flower & \ms{0.82}{0.03} & \ms{0.80}{0.02} & \ms{0.88}{0.07} & \msb{0.94}{0.01} & \ms{0.56}{0.06} & \ms{0.54}{0.04} & \ms{0.78}{0.02} & \ms{0.76}{0.12} \\
 & UniVLA & \ms{0.88}{0.03} & \ms{0.85}{0.01} & \ms{0.81}{0.06} & \msb{0.90}{0.01} & \ms{0.82}{0.05} & \ms{0.86}{0.04} & \ms{0.77}{0.01} & \ms{0.88}{0.03} \\
 & GR-1 & \ms{0.89}{0.03} & \msb{0.92}{0.03} & \ms{0.91}{0.03} & \ms{0.87}{0.04} & \ms{0.80}{0.01} & \ms{0.79}{0.09} & \ms{0.83}{0.05} & \ms{0.85}{0.05} \\
\addlinespace[2pt]
ALOHA-transfer & ACT & \ms{0.53}{0.02} & \msb{0.80}{0.06} & \ms{0.56}{0.06} & \ms{0.75}{0.18} & \ms{0.77}{0.03} & \ms{0.78}{0.05} & \ms{0.74}{0.08} & \ms{0.64}{0.06} \\
\addlinespace[2pt]
ALOHA-insertion & ACT & \ms{0.53}{0.04} & \ms{0.53}{0.04} & \ms{0.52}{0.03} & \ms{0.50}{0.00} & \msb{0.60}{0.03} & \ms{0.57}{0.09} & \ms{0.51}{0.10} & \ms{0.50}{0.00} \\
\addlinespace[2pt]
ALOHA-real-cup & ACT & \ms{0.71}{0.06} & \ms{0.80}{0.08} & \ms{0.83}{0.01} & \ms{0.75}{0.06} & \ms{0.86}{0.06} & \ms{0.75}{0.05} & \msb{0.89}{0.02} & \ms{0.83}{0.01} \\
\addlinespace[2pt]
ALOHA-real-drawer & PI$_{0.5}$ & \ms{0.50}{0.13} & \ms{0.58}{0.12} & \ms{0.57}{0.12} & \ms{0.58}{0.13} & \ms{0.71}{0.04} & \ms{0.62}{0.01} & \ms{0.69}{0.00} & \msb{0.76}{0.02} \\
\multicolumn{2}{l}{\textbf{Average}} & 0.72 & 0.76 & 0.74 & \textbf{0.77} & 0.76 & 0.73 & \textbf{0.77} & \textbf{0.77} \\
\bottomrule
\end{tabular}%
}
\vspace{-1.5em}
\end{table*}

\section{Experiment Setup}

\noindent\textbf{Benchmarks.} LIBERO~\citep{liu2023libero} (the LIBERO-10 subset of 10 long-horizon manipulation tasks); CALVIN~\citep{mees2022calvin} (each subtask in a chained sequence treated as an independent success/failure instance); and ALOHA~\citep{zhao2023learning}, spanning simulated (transfer, insertion) and real-world tasks, where we collect 120 cup-placement, 60 pick-cube, and 75 drawer-open-block-placement trajectories on the physical platform.
We also test the robustness of our method on policies trained with noisy
demonstrations from Layman V1.0~\citep{10802106}.

\noindent\textbf{Models.} We evaluate six state-of-the-art VLAs: PI$_0$~\citep{black2024pi_0} and PI$_{0.5}$~\citep{Intelligence202505AV} on LIBERO-10; FLOWER~\citep{reuss2025flower}, UNIVLA~\citep{bu2025univla}, and GR-1~\citep{wu2024unleashing} on CALVIN; and ACT~\citep{zhao2023learning} and PI$_{0.5}$ on ALOHA, collecting both successful and failed trajectories for each.

\noindent\textbf{Evaluation.} We aggregate per-timestep scores by max-pooling, $P^{\xi}_{\max} = \max_{t} P^{\xi}_{t}(\mathrm{Failure}\mid x_t)$, 
mirroring online deployment where an alarm fires once the score crosses $\theta(t)$. Because successful and failed rollouts end at different lengths, we truncate all trajectories to the shortest successful one before pooling, and report AUC on $P^{\xi}_{\max}$. We also report balanced accuracy (BA) over trajectory progress.

\noindent\textbf{Baselines.} 
We compare against \emph{embedding-based} detectors that classify internal model representations (SAFE~\citep{Gu2025SAFEMF}, Fail-Detect~\citep{xu2025can}, FIPER~\citep{romer2025failure}) and the \emph{score-based} STAC~\citep{agia2024unpacking}, as well as its variant within our full GRU pipeline (denoted GRU‑STAC).
To isolate the contribution of MI itself, we also re-implement four alternative action-continuity measures (variance, KNN distance, dot-product similarity, temporal gradient) within our GRU framework. Dataset statistics, training and threshold-calibration details, and these alternatives are in Appendix~\ref{app:Experiment-Details}.

\begin{figure}
    \centering
    \includegraphics[width=\linewidth]{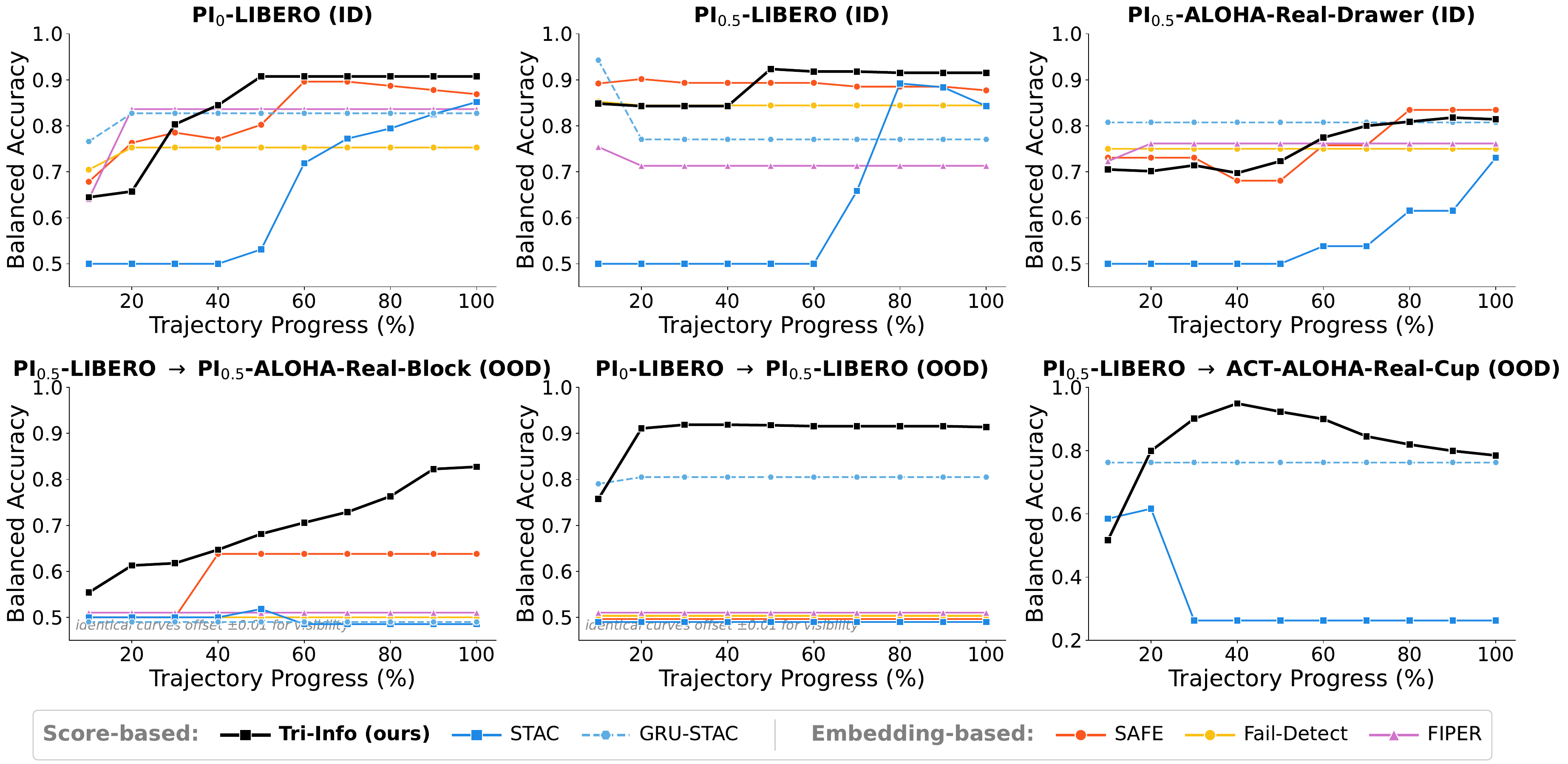}
    \caption{\textbf{Baseline comparison.} Failure detection accuracy
across different trajectory progress. Our Tri-Info detector
achieves competitive accuracy at both early and final time points, and transfers to OOD settings, where no baseline holds up consistently. Embedding-based baselines are omitted on
PI$_{0.5}$-LIBERO $\rightarrow$ ACT-ALOHA-real-cup, where latent dimensions do
not match (a VAE-matched variant is evaluated in Appendix~\ref{app:Experiment-Details}).}
    \label{fig:baseline_accuracy_results}
    \vspace{-1.5em}
\end{figure}

\section{Results}
\label{sec:results}

\subsection{Predictive Power of Information-Theoretic Metrics}
\label{sec:single_metric}

\noindent\textbf{Every metric carries signal; temporal modeling matters.}
We feed each of the eight metrics as the sole input to two detectors: a
memoryless logistic regression (LogReg) that scores it instantaneously, and a
GRU that models its temporal evolution (Table~\ref{tab:single_feature}). Under
LogReg, all eight reach $\geq 0.65$ average in-domain mBA, with the strongest,
$\mathbb{H}(\mathbf{A}_t)$, at $0.75$, confirming that the predictive signal resides in
the metrics themselves. Modeling temporal context with a GRU lifts all metrics
to $0.72$--$0.77$, and the weakest metrics gain the most, so the gap between the
best and worst metric closes from $0.10$ to $0.05$. 
In conclusion, the metrics are individually informative, and modeling their temporal evolution improves detection performance.

\noindent\textbf{Qualitative behavior of the Tri-Info detector.}
On a representative ALOHA-real trajectory (Figure~\ref{fig:failure_trajectory_demo}), the Tri-Info detector raises $P(\mathrm{Failure})$ just \emph{before} each of the two windows ($t{=}97$--$117$ and $t{=}390$--$477$) where the cup nearly tips or falls, demonstrating early warning ahead of the visible danger. Figure~\ref{fig:suc_fail_score} corroborates this on LIBERO-10 (PI$_{0.5}$): successful trajectories stay below the threshold throughout, whereas failed ones cross it.
\begin{wrapfigure}{r}{0.42\textwidth}
\centering
\includegraphics[width=\linewidth]{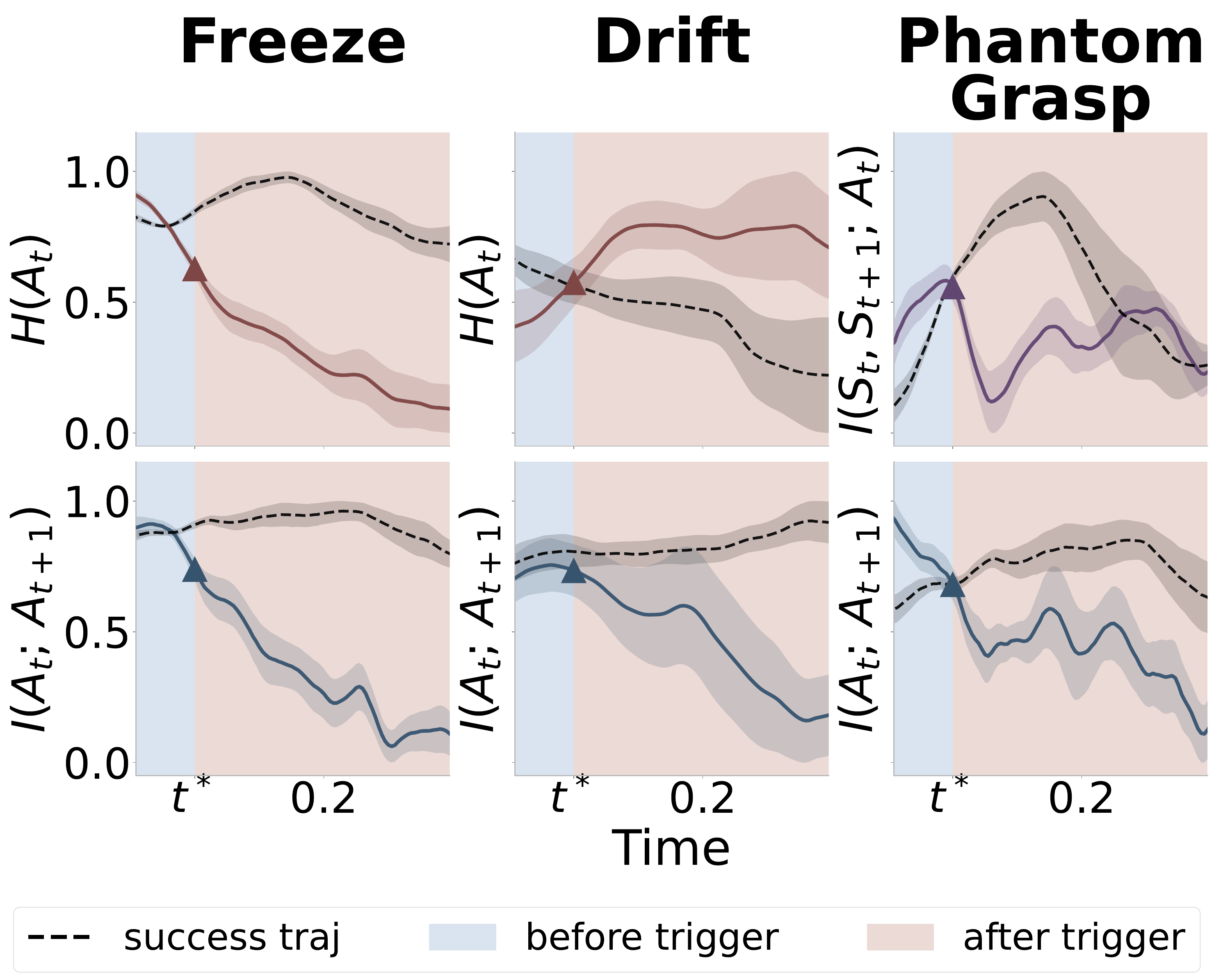}
\caption{\textbf{Failure signatures at the alarm.} Annotated failures ($n{=}20$
per mode) aligned at the detector's trigger $t^{*}$, against successful
trajectories at matched progress (black). Time is the lag from the trigger as a
fraction of the episode horizon, $(t-t^{*})/T$. The metric trends align with the
semantic meaning of each failure.}
\label{fig:interp}
\vspace{-2em}
\end{wrapfigure}

\noindent\textbf{Interpretability: the metrics align with failure semantics.}
We annotate failed rollouts as freeze, drift, or phantom grasp ($n{=}20$ per
mode) and examine each mode's two signature metrics around the trigger $t^{*}$
along two axes: the change from the trajectory's own pre-trigger level, and the
gap to successful trajectories at matched progress (Figure~\ref{fig:interp}).
The two axes agree in all six panels. The mutual information
$\mathbb{I}(\mathbf{A}_t;\mathbf{A}_{t+1})$ falls in every mode while that of
successful trajectories holds or rises: consecutive commands no longer predict
one another. The second metric separates the modes. In freeze,
$\mathbb{H}(\mathbf{A}_t)$ also collapses as the arm issues less varied actions.
In drift, it surges compared with successful trajectories, reflecting
erratic motion. In phantom grasp,
$\mathbb{I}(\mathbf{S}_t,\mathbf{S}_{t+1};\mathbf{A}_t)$ steps down while that
of successful trajectories rises, as the arm repeats a grasp that no longer
moves the objects.

\begin{wrapfigure}{r}{0.42\textwidth}
\vspace{-1.5em}
\centering
\includegraphics[width=\linewidth]{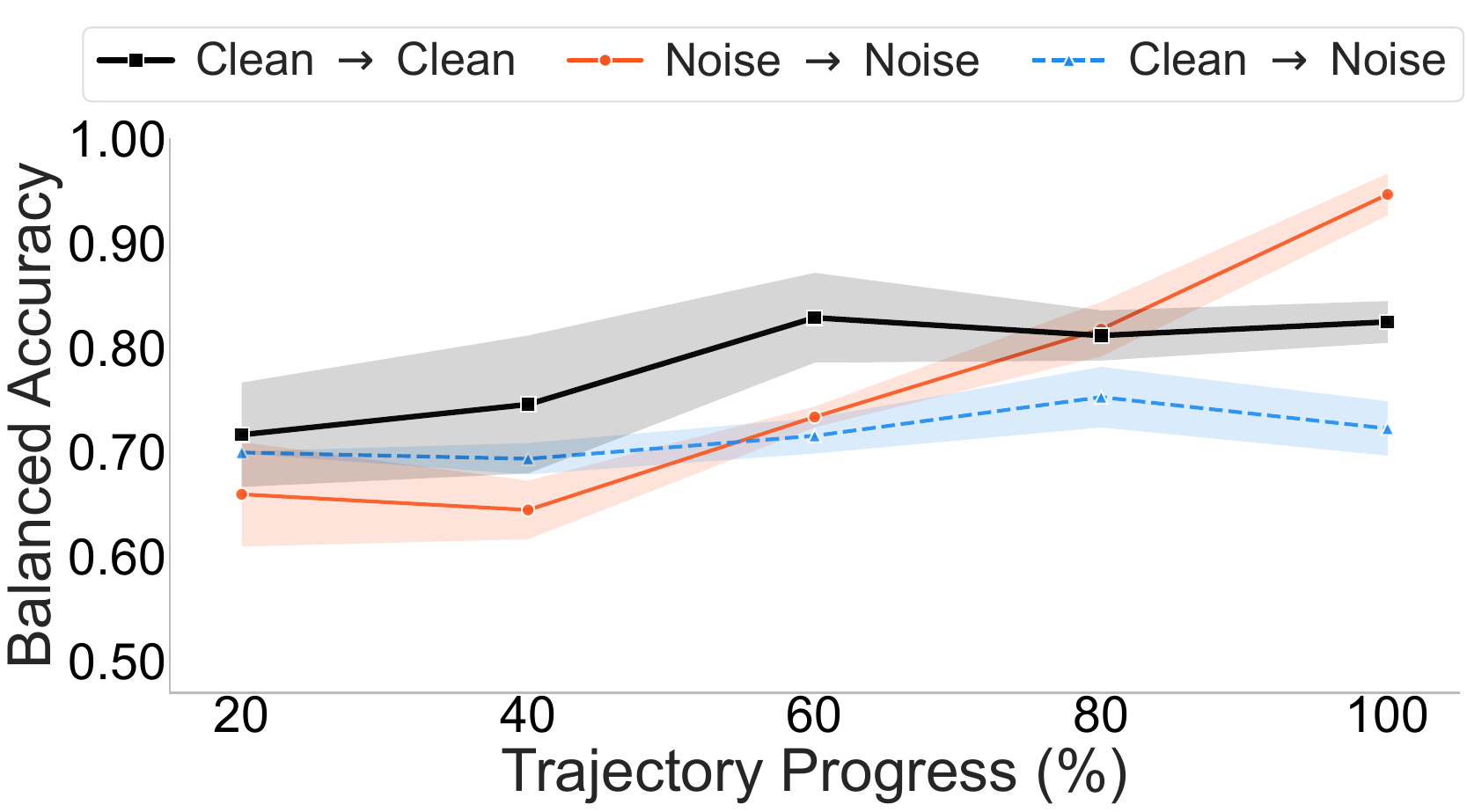}
\caption{\textbf{Tri-Info detection BA $\pm$ std across training-data sources under clean and noisy conditions.}
Tri-Info remains accurate on a policy trained on noisy demonstrations, and can transfer from clean-policy rollouts to noisy-policy rollouts.}
\label{fig:noisy}
\vspace{-1.2em}
\end{wrapfigure}

\noindent
\textbf{Robustness to policies trained on non-expert demonstrations.}
Real-world data is often collected by non-experts and therefore contains noisy,
indirect trajectories that nonetheless reach the goal. We thus test whether our detector remains reliable for a policy fine-tuned on such noisy data, and
can a detector trained only on clean-policy rollouts transfer to a noisy-policy one?
We use Layman V1.0~\citep{10802106}, which provides non-expert demonstrations
and a 20\%-spoiled \emph{Can} split: \textbf{Clean} (120 expert demos, 277 steps) vs.\
\textbf{Noisy} (120 expert + 30 lay-user demos in which operators
struggle to control the robot but eventually complete the task, 503 steps).
We fine-tune PI$_{0.5}$  on each pool, obtaining
SR 0.670 (clean) and 0.700 (noisy), and sample rollouts from both.
We report BA at each \% of trajectory progress in Figure \ref{fig:noisy} with three seeds each. The detector remains reliable on the noisy-data policy (mBA 0.728), and a detector
trained \emph{only} on clean-policy rollouts can transfer to the noisy policy with good performance (mBA 0.716).

\subsection{Baseline Comparison}
\label{sec:baseline}
We compare the Tri-Info detector against embedding-based methods
(SAFE~\citep{Gu2025SAFEMF}, Fail-Detect~\citep{xu2025can},
FIPER~\citep{romer2025failure}), the score-based STAC~\citep{agia2024unpacking},
and its GRU variant in our pipeline (GRU-STAC). Beyond the in-domain setting, we
evaluate three transfers of increasing severity: cross-model, cross-environment, and cross-model-environment. Because embedding-based latents are
architecture-specific, the embedding-based detectors cannot be applied in the
last setting, where the target policy's latent dimensionality differs from the
source. Figure~\ref{fig:baseline_accuracy_results} reports balanced accuracy
versus trajectory progress.

\noindent\textbf{In-domain: Tri-Info is competitive with existing detectors.}
Tri-Info averages $0.83$ balanced accuracy (BA) in domain, against $0.82$ for
SAFE and $0.81$ for GRU-STAC. It reaches $0.84$ and $0.89$ on PI$_0$- and
PI$_{0.5}$-LIBERO-10, where SAFE reaches $0.82$ and $0.89$, and $0.76$ on
PI$_{0.5}$-ALOHA-real-drawer, where GRU-STAC is highest at $0.81$. Raw STAC is
the only method below $0.70$ ($0.61$), which shows that the consistency
statistic needs the temporal model to be predictive. All remaining methods fall
within $0.07$ BA of each other, so in-domain accuracy does not distinguish
them. The distinguishing question is whether that accuracy survives a change of
model or embodiment.

\noindent\textbf{Cross-domain: only Tri-Info retains accuracy under shift.}
We evaluate three transfers of increasing severity in Figure \ref{fig:baseline_accuracy_results}. Changing the policy alone
(PI$_0$-LIBERO $\rightarrow$ PI$_{0.5}$-LIBERO) already separates the methods:
\begin{wraptable}{r}{0.42\textwidth}
\caption{Multi-metric detector ablation on mean balanced accuracy. ID is pooled in-domain; OOD averages cross-model and sim-to-real transfer.}
\label{tab:metric_ablation}
\centering
\scriptsize
\setlength{\tabcolsep}{3pt}
\renewcommand{\arraystretch}{1.0}
\vspace{-1.5em}
\begin{tabular}{@{}ccc cc@{}}
\toprule
$\mathbb{H}(\mathbf{A}_t)$ & $\mathbb{I}(\mathbf{A}_t;\mathbf{A}_{t+1})$ & $\mathbb{I}(\mathbf{S}_t,\mathbf{S}_{t+1};\mathbf{A}_t)$ & ID & OOD \\
\midrule
\rowcolor{bestrow}
\checkmark & \checkmark & \checkmark & \best{.88} & \best{.87} \\
\midrule
\checkmark & \checkmark &            & \best{.87} & \best{.85} \\
\checkmark &            & \checkmark & .86        & .70        \\
           & \checkmark & \checkmark & .86        & .84        \\
\midrule
\checkmark &            &            & \best{.87} & .60        \\
           & \checkmark &            & .84        & \best{.77} \\
           &            & \checkmark & \best{.82} & .67        \\
\midrule
\multicolumn{2}{@{}l}{All 8 Features}  && .87 & .74 \\
\bottomrule
\end{tabular}
\vspace{-1em}
\end{wraptable}
Tri-Info reaches $0.90$ and GRU-STAC $0.80$, while SAFE, Fail-Detect, and FIPER
fall to $0.50$, because their classifiers are fitted to the absolute geometry of
one policy's latent space. Changing the environment instead
($\rightarrow$ PI$_{0.5}$-ALOHA-real-block) leaves Tri-Info at $0.70$ and SAFE
at $0.60$, with all other baselines at $0.50$. Changing model, embodiment,
action dimensionality, and visual appearance at once
($\rightarrow$ ACT-ALOHA-real-cup) makes the embedding-based detectors
inapplicable, since their input dimensionality is fixed to the source policy;
GRU-STAC reaches $0.76$ and STAC falls to $0.33$, below chance, so its score
anti-correlates with failure, whereas Tri-Info attains $0.82$. A VAE
projection that matches latent dimensions restores partial applicability for the
embedding-based detectors
(Appendix~\ref{app:Experiment-Details}).

\begin{table}[t]
\centering
\caption[Estimator-input ablation]{Estimator-input ablation. Mean balanced
accuracy over $t\in\{0.1,\ldots,1.0\}$, three seeds. Splits and block means follow
Table~\ref{tab:metric_ablation}, with 
cross-model being PI$_{0}$-LIBERO to PI$_{0.5}$-LIBERO and
sim-to-real being PI$_{0.5}$-LIBERO to
ACT-ALOHA-real. $\mathbf{a}_t,\mathbf{s}_t$ are the VAE latents,
$\mathbf{s}_t$ fusing image, wrist-image and proprioceptive features; $a_t$ is
the executed action and $q_t$ the raw proprioception, both used without a VAE.}
\label{tab:input_ablation}
\small
\setlength{\tabcolsep}{6pt}
\resizebox{0.99\textwidth}{!}{%
\begin{tabular}{@{}l ccc @{\hspace{1.6em}} ccc@{}}
\toprule
Inputs & PI$_0$-LIBERO & PI$_{0.5}$-LIBERO & ID AVG. & Cross-Model & Sim-to-Real & OOD AVG. \\
\midrule
$\mathbf{a}_t+\mathbf{s}_t$ (ours)
& \dcell{0.870}{} & \dcell{0.893}{} & \dcell{0.881}{}
& \dcell{0.906}{} & \dcell{0.824}{} & \dcell{\best{0.865}}{} \\
\midrule
$\mathbf{a}_t+q_t$
& \dcell{0.864}{\dloss{0.006}} & \dcell{0.888}{\dloss{0.005}} & \dcell{0.876}{\dloss{0.005}}
& \dcell{0.913}{\dgain{0.007}} & \dcell{0.547}{\dloss{0.277}} & \dcell{0.730}{\dloss{0.135}} \\
$a_t+\mathbf{s}_t$
& \dcell{0.907}{\dgain{0.037}} & \dcell{0.887}{\dloss{0.006}} & \dcell{\best{0.897}}{\dgain{0.016}}
& \dcell{0.771}{\dloss{0.135}} & \dcell{0.752}{\dloss{0.072}} & \dcell{0.761}{\dloss{0.104}} \\
$a_t+q_t$ (API-only)
& \dcell{0.878}{\dgain{0.008}} & \dcell{0.882}{\dloss{0.011}} & \dcell{0.880}{\dloss{0.001}}
& \dcell{0.859}{\dloss{0.047}} & \dcell{0.729}{\dloss{0.095}} & \dcell{0.794}{\dloss{0.071}} \\
\bottomrule
\end{tabular}
}
\vspace{-4mm}
\end{table}

\subsection{Ablation Study}
\label{sec:metric_ablation}

\noindent \textbf{Metric ablation: Tri-Info's three signals contribute to OOD performance.}
We compare the three-metric detector against all of its 2- and 1-metric subsets and the full 8-metric fusion (Table~\ref{tab:metric_ablation}). The three-metric set improves over every smaller subset on both ID (0.88) and OOD (0.87), while expanding to eight metrics holds ID flat (0.87) and substantially degrades OOD to 0.74 -- the additional five signals overfit in-distribution patterns and fail to transfer.

\noindent \textbf{Input ablation: how much of the signal needs model internals?}
Everything so far is estimated on VAE latents of the policy's internal
embeddings, which requires white-box access. We re-ran the pipeline changing
only the estimator inputs, replacing $\mathbf{a}_t$ by the executed action $a_t$
and $\mathbf{s}_t$ by the raw proprioception $q_t$, both visible from outside the
policy and low-dimensional enough to enter the estimators without a VAE, unlike
the raw observation $s_t$. In domain, the
internal representations make no difference to detection accuracy: the four
choices lie between 0.876 and 0.897, and the highest is $a_t+\mathbf{s}_t$.
Under shift they do: the OOD mean falls from 0.865 with the embeddings to 0.794
for the black-box $a_t+q_t$. Tri-Info therefore runs in a fully black-box
setting, and the white-box requirement earns its cost only under domain shift.

\section{Conclusion}
\vspace{-0.5em}
We formalized VLA control as a closed-loop information pipeline and reduced
eight derived metrics to three complementary Tri-Info metrics
$\{\mathbb{H}(\mathbf{A}_t),\ \mathbb{I}(\mathbf{A}_t; \mathbf{A}_{t+1}),\ \mathbb{I}(\mathbf{S}_t, \mathbf{S}_{t+1}; \mathbf{A}_t)\}$
that diagnose action collapse, temporal incoherence, and action--state decoupling.
Paired with a per-signal GRU detector, Tri-Info stays interpretable and
generalizable, transferring across VLA models, benchmarks, and the sim-to-real
gap to reach $70\%+$ on real-world tasks.

\bibliographystyle{plainnat}
\bibliography{references}

\newpage
\appendix
\section{Experiment Details}
\label{app:Experiment-Details}

We describe the datasets (Appendix~\ref{app:datasets}), the training and evaluation protocol (Appendix~\ref{app:protocol}), the baselines (Appendix~\ref{app:Baselines-Detail}), and additional results and sensitivity analyses (Appendix~\ref{app:additional-results}).

\begin{wrapfigure}{r}{0.5\textwidth}
\centering
\includegraphics[width=\linewidth]{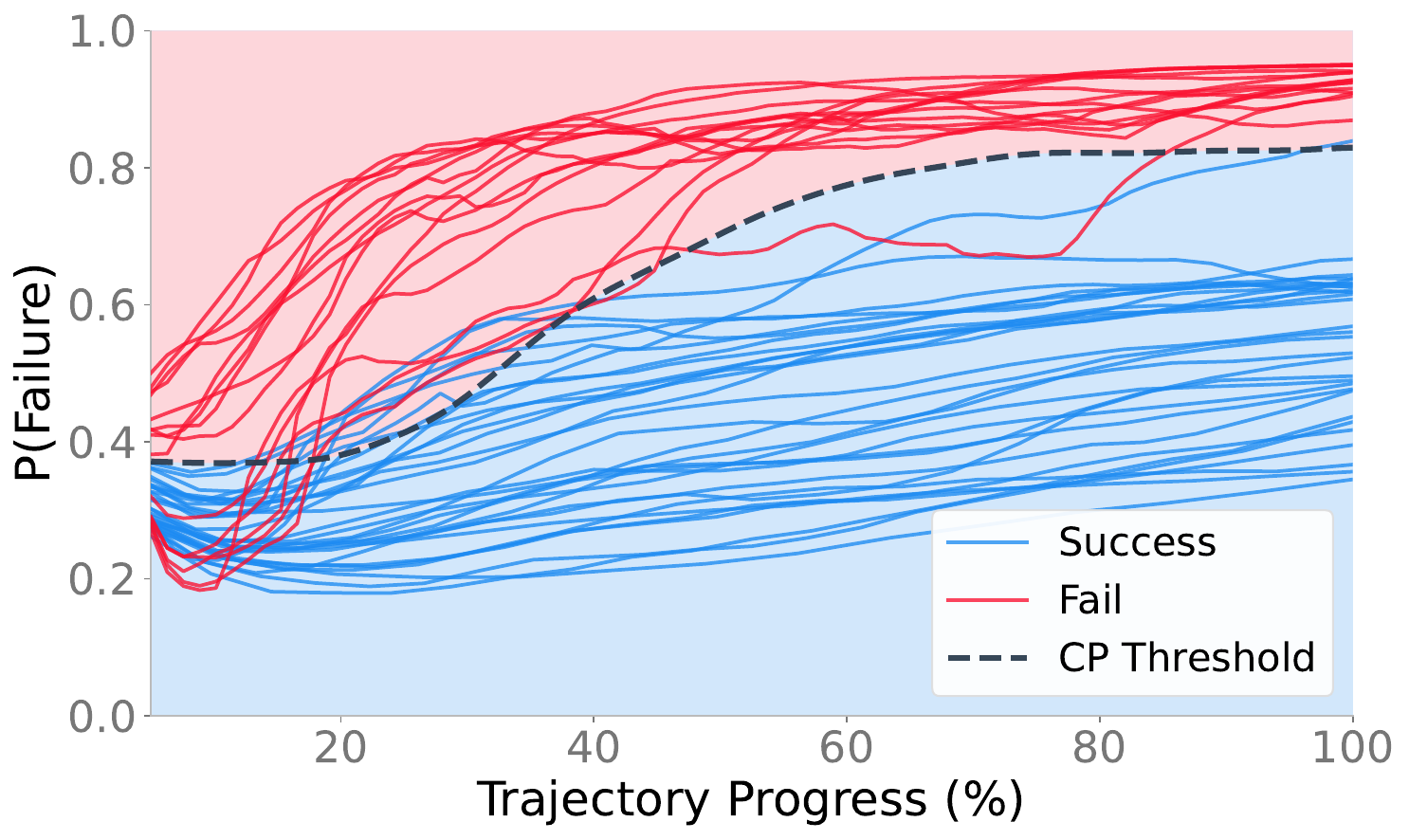}
\caption{\textbf{Tri-Info detector per-timestep $P(\mathrm{Failure})$ on PI$_{0.5}$-LIBERO over 60 random samples.} Failed trajectories (red) cross the threshold $\theta(t)$; successful ones (blue) stay below it.}
\label{fig:suc_fail_score}
\vspace{-2em}
\end{wrapfigure}

\subsection{Datasets}
\label{app:datasets}

Table~\ref{tab:model_data_statistics} summarizes the data collected for each model--environment combination.

\begin{table}[t]
\centering
\caption{Dataset statistics.}
\label{tab:model_data_statistics}
\begin{tabular}{lcccc}
\toprule
\textbf{Dataset} & \textbf{Success} & \textbf{Fail} & \textbf{Total} & \textbf{Suc\%} \\
\midrule
PI$_0$-LIBERO & 245 & 55 & 300 & 81.7\% \\
PI$_{0.5}$-LIBERO & 875 & 75 & 950 & 92.1\% \\
ACT-ALOHA-real & 60 & 60 & 120 & 50.0\% \\
PI$_{0.5}$-ALOHA-real-block& 26 & 34 & 60 & 43.3\% \\
PI$_{0.5}$-ALOHA-real-drawer& 31 & 44 & 75 & 41.3\% \\
FLOWER-CALVIN & 224 & 76 & 300 & 74.7\% \\
ACT-ALOHA-sim-insertion & 65 & 235 & 300 & 21.7\% \\
ACT-ALOHA-sim-transfer & 259 & 41 & 300 & 86.3\% \\
GR-1-CALVIN & 360 & 50 & 410 & 87.8\% \\
UniVLA-CALVIN & 215 & 35 & 250 & 86.0\% \\
\midrule
\textbf{Total} & \textbf{2360} & \textbf{705} & \textbf{3065} & \textbf{77.0\%} \\
\bottomrule
\end{tabular}
\end{table}

\vspace{0.5em}
\noindent\textbf{Real-world data collection.}
Real-world demonstrations were collected on the ALOHA platform (AgileX {cobot\_magic} variant) using a single 6-DoF arm with a parallel gripper and three RGB cameras (one global and two wrist-mounted). We recorded demonstrations via master--puppet teleoperation after sensor synchronization. Each trajectory logs the arm's 7-dimensional joint positions (six joints plus the gripper), the master-side control commands, and synchronized $480\times640$ RGB observations, all sampled at $50\,\mathrm{Hz}$.

\begin{figure}
    \centering
    \includegraphics[width=\linewidth]{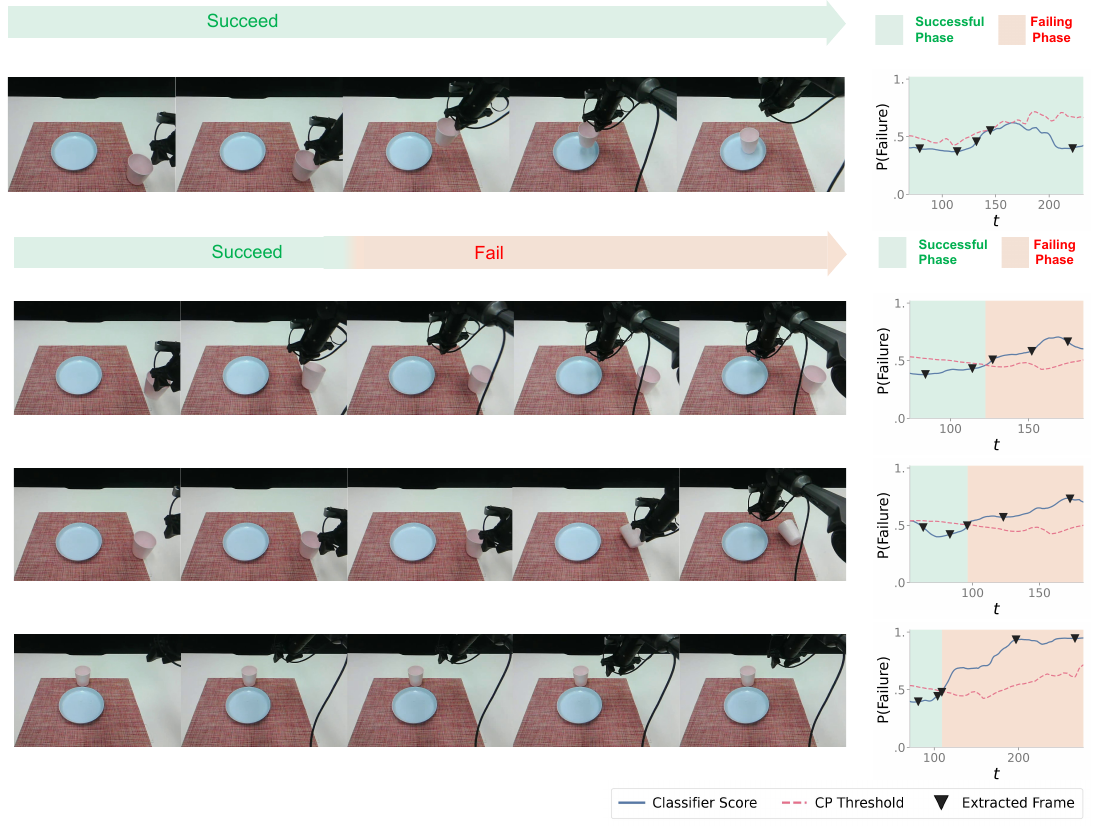}
    \includegraphics[width=\linewidth]{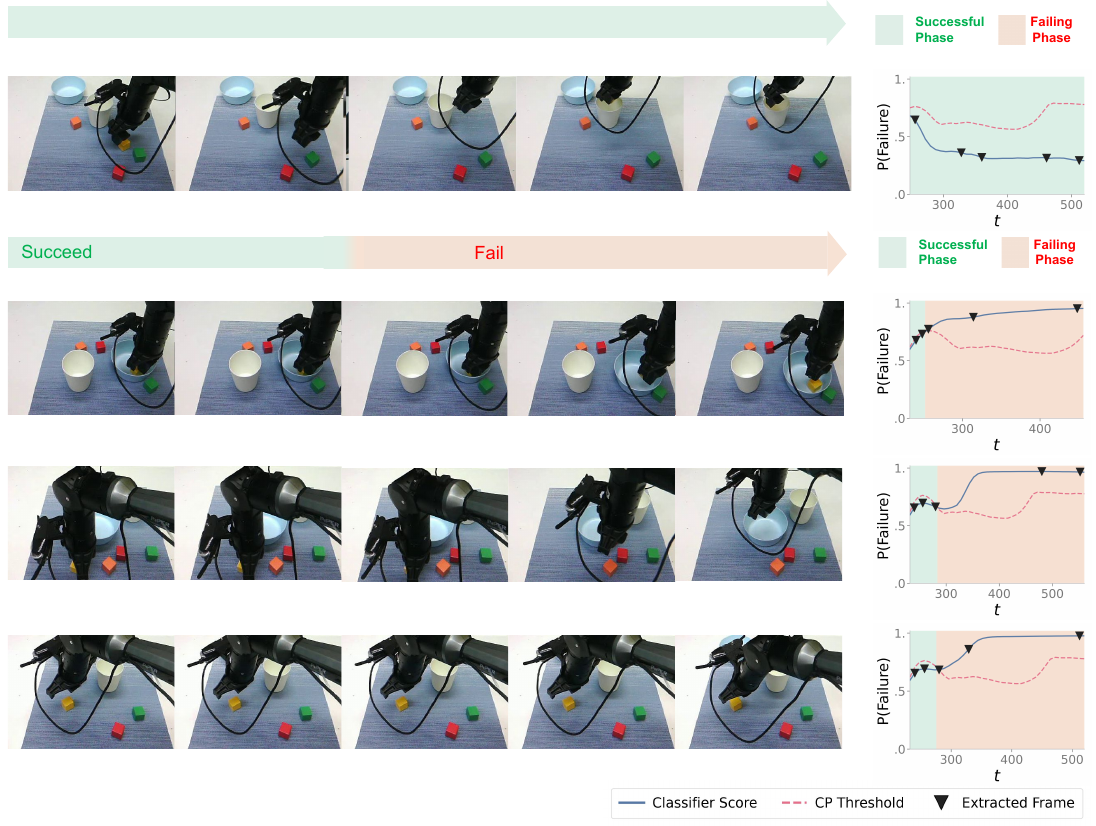}
    \caption{Tri-Info detector results for two real-world platforms (top: ACT-ALOHA-real; bottom: PI$_{0.5}$-ALOHA-real-block). Each platform shows four rollouts -- three failed and one successful -- and each rollout pairs five trajectory frames with the corresponding classifier score and CP threshold ($\alpha = 0.05$).}
    \label{fig:aloha-more-trajectories}
\end{figure}

\vspace{0.5em}
\noindent\textbf{Real-world task 1: Cup on plate (ACT).}
The task is to place a red cup onto a small blue plate. A trial succeeds if the cup comes to rest within the plate boundary without toppling; all other outcomes count as failures. We collected 100 teleoperated demonstrations, combined them with simulated data, and trained ACT with batch size 4 for 10{,}000 epochs. For evaluation, we collected 120 rollout trajectories from the trained policy. More trajectories with Tri-Info detector score are shown in Figure~\ref{fig:aloha-more-trajectories}.

\vspace{0.5em}
\noindent\textbf{Real-world task 2: Block in cup (PI$_{0.5}$).}
The task is to place a yellow block into a cup. A trial succeeds if the block comes to rest inside the cup; missed grasps, releases above the rim, and toppled cups; or put into other containers all count as failures. We collected 50 teleoperated demonstrations (about 400 frames each), converted them to the LeRobot v2.0 format, and fine-tuned the pretrained PI$_{0.5}$ base model with \texttt{openpi} for 100{,}000 steps (global batch size 64, cosine schedule peaking at $5\times10^{-5}$), reusing the Trossen normalization statistics. For evaluation, we collected 60 rollout trajectories from the trained policy. Trajectories with Tri-Info detector score are shown in Figure~\ref{fig:aloha-more-trajectories}.

\begin{figure}
    \centering
    \includegraphics[width=\linewidth]{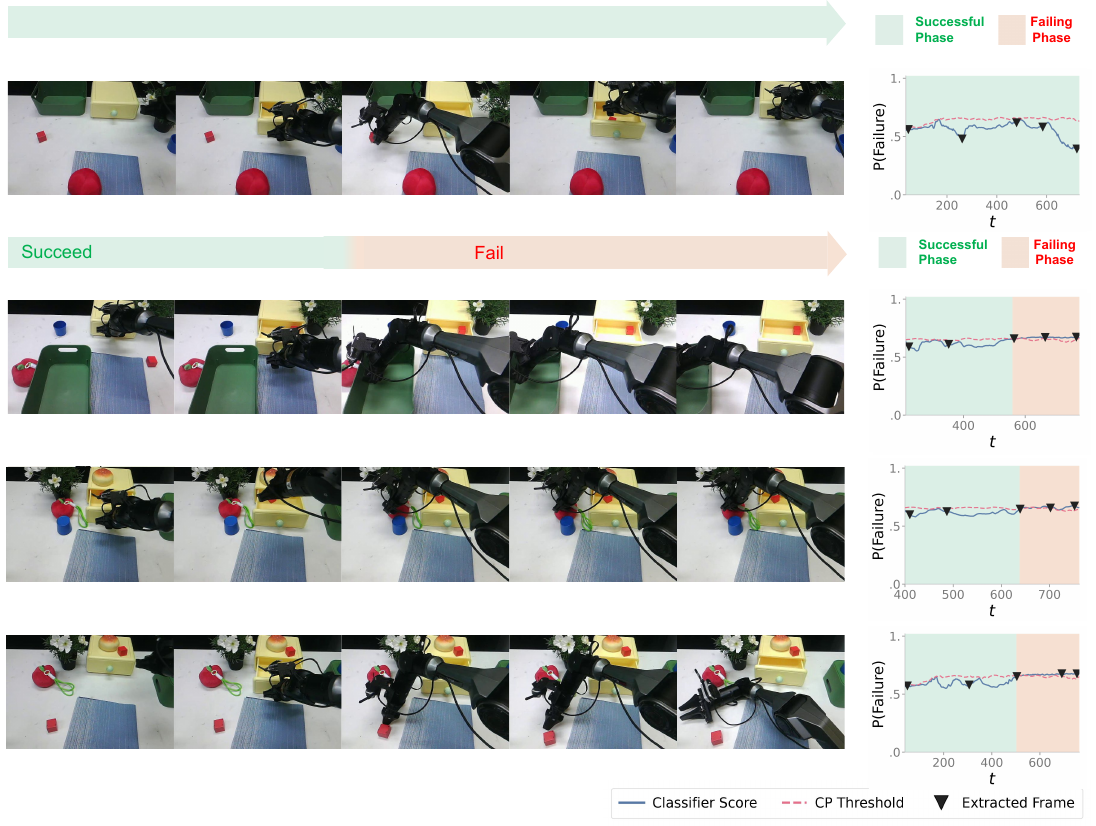}
    \caption{Tri-Info detector results for  PI$_{0.5}$-ALOHA-real-drawer. It shows four rollouts -- three failed and one successful -- and each rollout pairs five trajectory frames with the corresponding classifier score and CP threshold ($\alpha = 0.05$).}
    \label{fig:aloha-more-trajectories-drawer}
\end{figure}

\vspace{0.5em}
\noindent\textbf{Real-world task 3: Block into drawer (PI$_{0.5}$).}
The task is a four-stage sequence: open the drawer, pick up the red block, place it
inside, and close the drawer. A trial counts as a failure if the block is not placed
inside the drawer, or if the drawer cannot be closed afterwards. We collected 56
teleoperated demonstrations of roughly 650 frames each, converted them to the LeRobot
v2.0 format, and fine-tuned the pretrained PI$_{0.5}$ base model with \texttt{openpi}
for 30{,}000 steps (global batch size 32, cosine schedule peaking at $2.5\times10^{-5}$);
normalization statistics were computed from the same 56 demonstrations. For evaluation
we collected 75 rollout trajectories from the trained policy, of which 31 succeeded and
44 failed (41.3\% success rate). Trajectories with their Tri-Info detector scores are
shown in Figure~\ref{fig:aloha-more-trajectories-drawer}.

\subsection{Training and Evaluation Protocol}
\label{app:protocol}

This section describes how data are split, how the detectors are trained, and where each component of the pipeline is fitted.

\vspace{0.5em}
\noindent\textbf{Trajectory-level splits.}
Every split is made over whole trajectories, \emph{before} any sliding window is
formed, and is stratified by the outcome label $f^\xi$. A window $\mathbf{W}_t$
therefore never spans two trajectories, and no trajectory contributes windows to
more than one split, so near-duplicate windows cannot straddle the split
boundary. In-domain experiments split each dataset $70/30$ into a training and a
test split; transfer experiments fit everything on the source domain and
evaluate on target-domain trajectories the detector has never seen. The
calibration set $\mathcal{D}^{+}_{\mathrm{cal}}$ is the success subset of the \emph{source
training split}, so calibration consumes no target trajectory. Because those
successes also train the GRUs, $\theta(t)$ is calibrated in-sample, and we
therefore read $\alpha$ as an operating-point knob
(Section~\ref{sec:metric_ablation}) rather than an exact finite-sample coverage
guarantee.

\vspace{0.5em}
\noindent\textbf{Failure detector training.}
We train classifiers at the timestep level, where each timestep inherits the
trajectory-level label $f^{\xi}\in\{0,1\}$. The GRU model consists of a
single-layer GRU mapping the per-timestep metric scalar (input dimension one) to
a hidden state, followed by a two-layer MLP head that halves the hidden size
before a single output, with ReLU activation, dropout $0.1$, and a sigmoid
output. It is trained with Adam (learning rate $10^{-3}$, weight decay
$10^{-5}$); the hidden size and the number of epochs are not fixed but selected
per domain by the search of Appendix~\ref{app:hyperparameters}
(Table~\ref{tab:hp}). The logistic regression baseline uses a single linear layer
with sigmoid activation, trained with the L-BFGS solver at an inverse
regularization strength of $1.0$. Both models receive the per-timestep metrics,
normalized using global statistics computed across the entire training set.
Unless otherwise noted, every experiment is run with three random seeds, and we
report the average across seeds.

\vspace{0.5em}
\noindent\textbf{VAE training data and leakage.}
The projection of Appendix~\ref{app:mi_estimation} is fitted once per
model--dataset domain, on all of that domain's per-step embeddings -- test
trajectories included -- using reconstruction plus a weak KL penalty. A target domain therefore
contributes only unlabelled embeddings to its own projection, and no failure label enters the VAE training pipeline anywhere. The projection does stay
transductive in the unlabelled data; fitting it on a disjoint unlabelled pool
would remove even that.

\vspace{0.5em}
\noindent\textbf{Where each component is fitted.}
Table~\ref{tab:protocol} summarizes where each component is fitted. Only the
label-free front end is refitted for a new model or embodiment; the GRU weights,
the normalization constants $(\mu,\sigma)$, and the threshold $\theta(t)$ are
fitted once on the source training split and applied unchanged, with target
sequences z-normalized by \emph{source} statistics rather than their own. 

\begin{table}[h]
\centering
\caption{Where each component of the pipeline is
fitted. Only the label-free front end is refitted for a new model or embodiment.}
\label{tab:protocol}
\small
\setlength{\tabcolsep}{5pt}
\begin{tabular}{@{}llcc@{}}
\toprule
\textbf{Component} & \textbf{Fitted on} & \textbf{Uses label} & \textbf{Refit per domain} \\
\midrule
VAE $E_{\mathbf{S}},E_{\mathbf{A}}$ & unlabelled rollouts of that domain     & no  & yes \\
Normalization $(\mu,\sigma)$        & source training split                  & no  & no  \\
Per-metric GRUs                     & source training split                  & yes & no  \\
Threshold $\theta(t)$               & successes of the source training split & yes & no  \\
\bottomrule
\end{tabular}
\end{table}

\subsection{Baselines}
\label{app:Baselines-Detail}
This section describes the baseline failure detection methods used in our comparisons, as well as the alternative action continuity metrics we evaluate to validate our design choices.
We compare against four recent failure detection methods, which we categorize based on their \emph{input modality into the classifier}: \textit{embedding-based} methods that use model representations, and \textit{score-based} methods that operate on computed scalar signals. This distinction is important, as embedding-based approaches are inherently tied to specific model architectures and cannot generalize across different policies without retraining.

\vspace{0.5em}
\noindent\textbf{Embedding-based methods.}
\textbf{SAFE}~\citep{Gu2025SAFEMF} trains a classifier on VLA latent representations to distinguish successful from failed executions, so both its input dimensionality and its decision geometry are fixed to that of embedding space, and it is bound to the architecture by construction.
\textbf{Fail-Detect}~\citep{xu2025can} distills the policy's inputs and outputs into a scalar signal, which makes it look model-agnostic. But its strongest score (logpZO) is a density estimator trained on one policy's observation embeddings. On a different model or task, the embeddings no longer match what it learned, so almost every trajectory looks out-of-distribution; the score saturates and can no longer tell success from failure. 
\textbf{FIPER}~\citep{romer2025failure} combines two signals: (i) out-of-distribution observations via Random Network Distillation in the policy's embedding space (RND-OE), and (ii) action-chunk entropy (ACE) over sampled actions. The RND-OE branch detects novelty using a predictor network trained on one policy's embeddings, so it is tied to that embedding space and fails once the model or task changes, for the same reason as Fail-Detect. The ACE branch, however, reads only the sampled action chunks and trains no model bound to the source policy, so it does carry over. We therefore take ACE as FIPER's transferable component and compare against it in Figure~\ref{fig:appendix_action_scores}.

\vspace{0.5em}
\noindent\textbf{Score-based methods.}
\textbf{STAC}~\citep{agia2024unpacking} computes the statistical divergence between overlapping segments of sampled action chunks to measure temporal consistency in generative policies. The original method requires generating multiple samples per timestep; 
our $\mathbb{I}(\mathbf{A}_t;\mathbf{A}_{t+1})$ shares a similar spirit but differs in two respects: (i) it operates on executed actions over a sliding window, compatible with deterministic VLA inference and adding much less overhead, whereas STAC compares only an overlapping sub-window of 32+ sampled futures; (ii) the decomposition $\mathbb{I}(\mathbf{A}_t;\mathbf{A}_{t+1}) = \mathbb{H}(\mathbf{A}_{t+1}) - \mathbb{H}(\mathbf{A}_{t+1}|\mathbf{A}_t)$ captures both diversity and consistency, while STAC measures consistency only.

\begin{figure}[t]
\centering
\includegraphics[width=\linewidth]{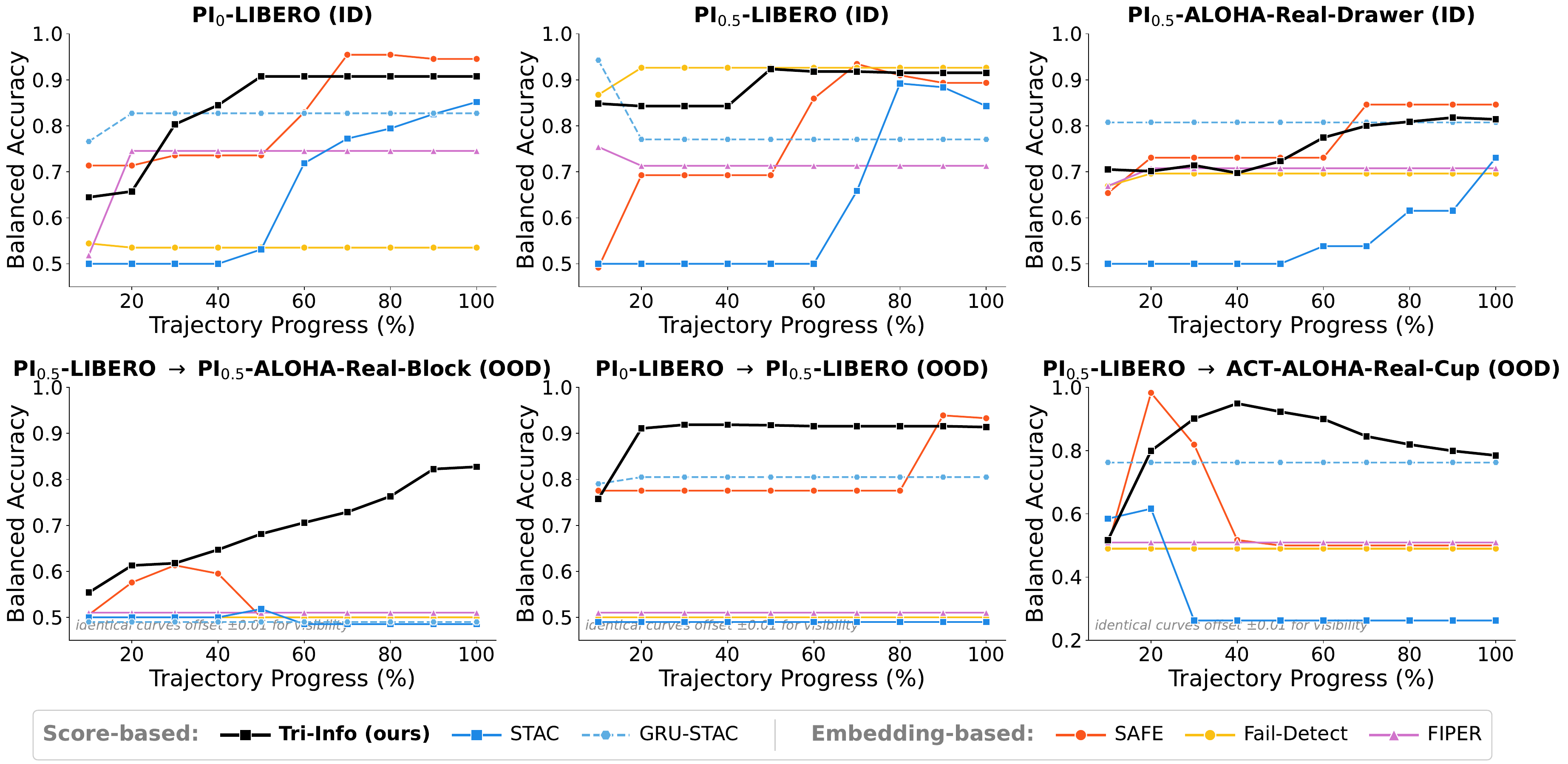}
\caption{\textbf{Dimension-matched baselines.} Balanced accuracy versus
trajectory progress for the three in-domain settings (top) and the three
transfers (bottom). All embedding-based detectors use the VAE-matched latent, so
they are applicable in every panel, including
PI$_{0.5}$-LIBERO $\rightarrow$ ACT-ALOHA-real-cup. In the OOD panels overlapping curves are offset by $\pm0.01$ for visibility.}
\label{fig:appendix_vae_baseline}
\end{figure}

\begin{figure}[p]
    \centering
    \includegraphics[width=\linewidth]{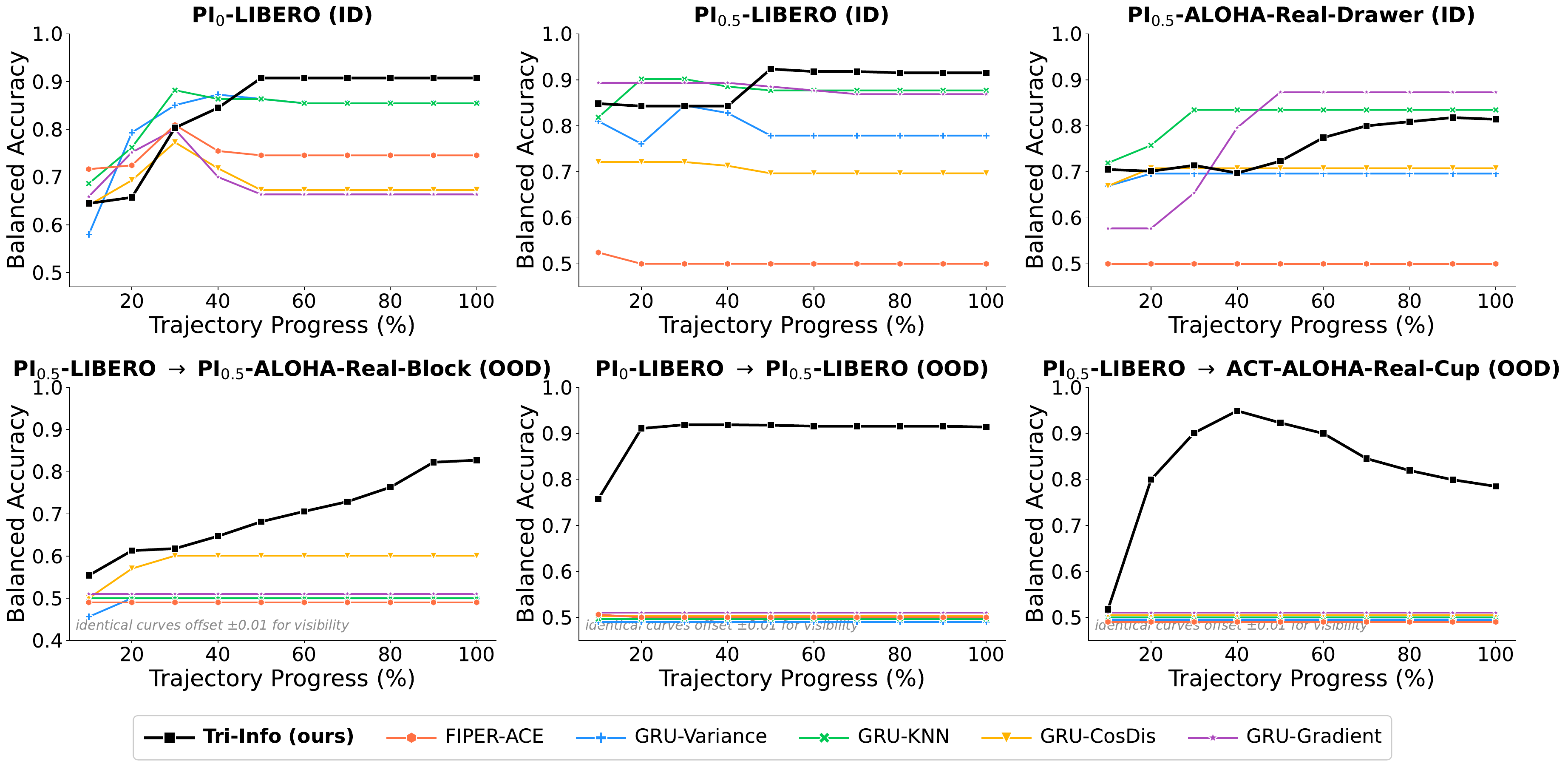}\\[0.8em]
    \includegraphics[width=\linewidth]{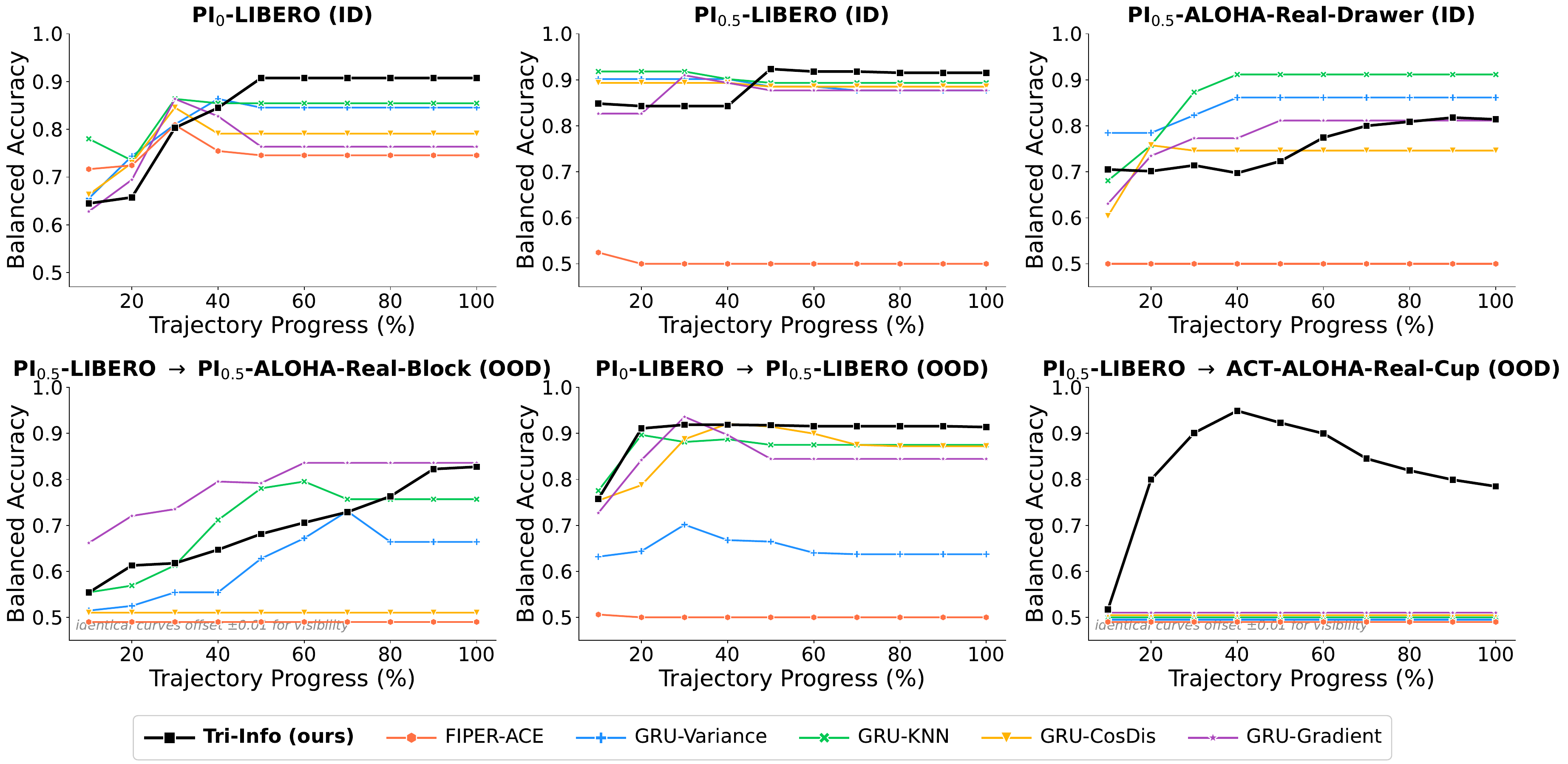}
    \caption{\textbf{Comparison against action-based scores.} Balanced accuracy
versus trajectory progress for Tri-Info, FIPER-ACE, and the four
action-continuity scores, all sharing our GRU pipeline. \textbf{Top:} scores
computed on the raw action embeddings, where all four at or near chance in
every OOD panel. \textbf{Bottom:} the same scores on the VAE-matched latent,
which recovers accuracy on the cross-environment and cross-model transfers but
not on PI$_{0.5}$-LIBERO $\rightarrow$ ACT-ALOHA-real-cup. Flat curves at chance
are offset by $\pm0.01$ for visibility.}
    \label{fig:appendix_action_scores}
\end{figure}

\vspace{0.5em}
\noindent\textbf{Dimension-matched embedding baselines with VAE compression.}
The embedding-based detectors cannot be applied when the target policy's latent
dimensionality differs from the source, which is the case for
PI$_{0.5}$-LIBERO $\rightarrow$ ACT-ALOHA-real-cup. To evaluate them in that
setting as well, we project the source and target embeddings through the
per-domain VAE to the common $d_z=7$ latent and train and apply each detector on
those latents, giving the variants SAFE-VAE, Fail-Detect-VAE, and FIPER-VAE.
Table~\ref{tab:full_baseline_results} reports trajectory-level AUC and mean
balanced accuracy for all methods, and Figure~\ref{fig:appendix_vae_baseline}
shows balanced accuracy versus trajectory progress for the dimension-matched
variants.

In domain, the projection lowers the average accuracy of all three
embedding-based detectors, as expected from compressing a high-dimensional
latent to seven dimensions: SAFE drops from $0.91$ to $0.85$ AUC and from $0.82$
to $0.79$ mBA, while Fail-Detect and FIPER lose at most $0.02$ AUC and $0.06$
mBA. Under shift, the shared latent makes cross-domain evaluation possible:
all three detectors now apply to
PI$_{0.5}$-LIBERO $\rightarrow$ ACT-ALOHA-real-cup, where the raw variants
cannot run. The shared latent also improves transfer for some detectors.
SAFE-VAE outperforms raw SAFE on both transfers where the two are comparable,
raising AUC from $0.61$ to $0.86$ on
PI$_{0.5}$-LIBERO $\rightarrow$ PI$_{0.5}$-ALOHA-real-block and from $0.64$ to
$0.74$ on PI$_0$-LIBERO $\rightarrow$ PI$_{0.5}$-LIBERO. On the latter transfer
it also raises mBA from $0.50$ to $0.81$. FIPER-VAE likewise raises AUC on
PI$_0$-LIBERO $\rightarrow$ PI$_{0.5}$-LIBERO from $0.56$ to $0.77$. The gain is
not uniform, however: Fail-Detect-VAE falls below chance on all three transfers
($0.32$ average AUC). Nor does it extend to deployable accuracy.
Fail-Detect-VAE and FIPER-VAE sit at $0.50$ mBA on every OOD setting, and
SAFE-VAE averages $0.64$. The cases where AUC stays high but mBA collapses
identify the failure mode. These cases are SAFE-VAE on the two transfers to real
ALOHA ($0.86$ vs.\ $0.53$ and $0.79$ vs.\ $0.58$) and FIPER-VAE on
PI$_0$-LIBERO $\rightarrow$ PI$_{0.5}$-LIBERO ($0.77$ vs.\ $0.50$). There the
scores still rank trajectories within the target domain, but their absolute
level shifts. The threshold $\theta(t)$ calibrated on the source domain
therefore no longer separates success from failure, and the deployed detector
operates at or near chance.

\begin{table}[t]
\centering\small
\setlength{\tabcolsep}{4pt}
\caption{Full baseline comparison. Trajectory-level AUC (top) and mean balanced
accuracy over $t\in\{0.1,\dots,1.0\}$ (bottom). ``VAE'' denotes the
dimension-matched variant of each embedding-based detector. ``n/a'' marks
settings where the raw detector cannot be applied because the target latent
dimensionality differs from the source. Best per row in bold.}
\label{tab:full_baseline_results}
\resizebox{0.99\textwidth}{!}{%
\begin{tabular}{lccccccccc}
\toprule
& \multicolumn{3}{c}{Score-based} & \multicolumn{6}{c}{Embedding-based} \\
\cmidrule(lr){2-4}\cmidrule(lr){5-10}
& & & & \multicolumn{2}{c}{SAFE} & \multicolumn{2}{c}{Fail-Detect} & \multicolumn{2}{c}{FIPER} \\
\cmidrule(lr){5-6}\cmidrule(lr){7-8}\cmidrule(lr){9-10}
Setting & Tri-Info & STAC & GRU-STAC & raw & VAE & raw & VAE & raw & VAE \\
\midrule
\multicolumn{10}{l}{\emph{AUC}}\\
PI$_0$-LIBERO-10 & 0.70 & 0.44 & 0.60 & \textbf{0.85} & 0.74 & 0.70 & 0.73 & 0.63 & 0.49 \\
PI$_{0.5}$-LIBERO-10 & 0.86 & 0.67 & 0.81 & \textbf{0.95} & 0.90 & 0.75 & 0.82 & 0.77 & 0.93 \\
PI$_{0.5}$-ALOHA-real-drawer & 0.90 & 0.75 & 0.85 & \textbf{0.94} & 0.90 & 0.89 & 0.72 & 0.81 & 0.75 \\
\textit{ID average} & 0.82 & 0.62 & 0.76 & \textbf{0.91} & 0.85 & 0.78 & 0.76 & 0.73 & 0.72 \\
\addlinespace
PI$_{0.5}$-LIBERO $\rightarrow$ PI$_{0.5}$-ALOHA-real-block & 0.81 & 0.65 & 0.77 & 0.61 & 0.86 & 0.32 & 0.24 & \textbf{0.91} & 0.51 \\
PI$_0$-LIBERO $\rightarrow$ PI$_{0.5}$-LIBERO & \textbf{0.82} & 0.41 & 0.52 & 0.64 & 0.74 & 0.69 & 0.49 & 0.56 & 0.77 \\
PI$_{0.5}$-LIBERO $\rightarrow$ ACT-ALOHA-real-cup & 0.89 & 0.27 & \textbf{0.94} & n/a & 0.79 & n/a & 0.22 & n/a & 0.50 \\
\textit{OOD average} & \textbf{0.84} & 0.44 & 0.74 & n/a & 0.80 & n/a & 0.32 & n/a & 0.59 \\
\midrule
\multicolumn{10}{l}{\emph{Mean balanced accuracy}}\\
PI$_0$-LIBERO-10 & \textbf{0.84} & 0.65 & 0.82 & 0.82 & 0.83 & 0.75 & 0.54 & 0.82 & 0.72 \\
PI$_{0.5}$-LIBERO-10 & 0.89 & 0.63 & 0.79 & 0.89 & 0.78 & 0.85 & \textbf{0.92} & 0.72 & 0.72 \\
PI$_{0.5}$-ALOHA-real-drawer & 0.76 & 0.55 & \textbf{0.81} & 0.76 & 0.77 & 0.75 & 0.69 & 0.76 & 0.70 \\
\textit{ID average} & \textbf{0.83} & 0.61 & 0.81 & 0.82 & 0.79 & 0.78 & 0.72 & 0.76 & 0.71 \\
\addlinespace
PI$_{0.5}$-LIBERO $\rightarrow$ PI$_{0.5}$-ALOHA-real-block & \textbf{0.70} & 0.49 & 0.50 & 0.60 & 0.53 & 0.50 & 0.50 & 0.50 & 0.50 \\
PI$_0$-LIBERO $\rightarrow$ PI$_{0.5}$-LIBERO & \textbf{0.90} & 0.50 & 0.80 & 0.50 & 0.81 & 0.50 & 0.50 & 0.50 & 0.50 \\
PI$_{0.5}$-LIBERO $\rightarrow$ ACT-ALOHA-real-cup & \textbf{0.82} & 0.33 & 0.76 & n/a & 0.58 & n/a & 0.50 & n/a & 0.50 \\
\textit{OOD average} & \textbf{0.81} & 0.44 & 0.69 & n/a & 0.64 & n/a & 0.50 & n/a & 0.50 \\
\bottomrule
\end{tabular}
}
\end{table}

\vspace{0.5em}
\noindent\textbf{Alternative action metrics.}
We further consider several metrics that summarize how an action sequence evolves
over time, each feeding into the same GRU-based detection pipeline. All are computed
over the same sliding window $\mathbf{W}_t=\{(\mathbf{s}_i,\mathbf{a}_i)\}_{i=t-W+1}^{t}$
of Section~\ref{sec:metrics}, using the action embeddings $\mathbf{a}_i$.
\begin{enumerate}
    \item \textbf{Variance} measures action dispersion within the window:
    $\text{Var}_t = \operatorname{tr}\operatorname{Cov}\big(\{\mathbf{a}_i\}_{i=t-W+1}^{t}\big)$.
    \item \textbf{KNN distance} measures local density:
    $\text{KNN}_t = \frac{1}{Wk}\sum_{i=t-W+1}^{t}\sum_{j=1}^{k}\|\mathbf{a}_i-\mathbf{a}_{\mathrm{nn}_j(i)}\|_2$,
    where $\mathrm{nn}_j(i)$ is the $j$-th nearest neighbor of $\mathbf{a}_i$ within $\mathbf{W}_t$.
    \item \textbf{Cosine dissimilarity} measures consecutive action alignment:
    $\text{CosDis}_t = 1 - \frac{1}{W-1}\sum_{i=t-W+1}^{t-1}\frac{\mathbf{a}_i^\top\mathbf{a}_{i+1}}{\|\mathbf{a}_i\|\,\|\mathbf{a}_{i+1}\|}$.
    \item \textbf{Temporal gradient} measures the rate of action change:
    $\text{Grad}_t = \frac{1}{W-1}\sum_{i=t-W+1}^{t-1}\|\mathbf{a}_{i+1}-\mathbf{a}_i\|_1$.
\end{enumerate}

\noindent These metrics summarize different aspects of an action sequence
(dispersion, local density, directional alignment, and rate of change). We
evaluate each one twice, on the raw action embeddings and on the VAE-matched
latent, which separates the contribution of
the information-theoretic form from that of the projection.

\begin{table}[t]
\centering\small
\setlength{\tabcolsep}{4pt}
\caption{Comparison against action-based scores, mean balanced accuracy over
$t\in\{0.1,\dots,1.0\}$. All scores share the GRU pipeline, so only the input
signal differs. ``Raw'' uses the action embeddings directly; ``VAE-matched''
uses the $d_z=7$ latent, which Tri-Info also
uses. FIPER-ACE reads sampled action chunks rather than embeddings and is
therefore unaffected by the projection. Best per row in bold.}
\label{tab:action_score_results}
\resizebox{0.99\textwidth}{!}{%
\begin{tabular}{lcccccccccc}
\toprule
& & & \multicolumn{4}{c}{Raw embeddings} & \multicolumn{4}{c}{VAE-matched latent} \\
\cmidrule(lr){4-7}\cmidrule(lr){8-11}
Setting & Tri-Info & FIPER-ACE & Var & kNN & CosDis & Grad & Var & kNN & CosDis & Grad \\
\midrule
PI$_0$-LIBERO-10 & \textbf{0.84} & 0.75 & 0.82 & 0.83 & 0.69 & 0.69 & 0.81 & \textbf{0.84} & 0.78 & 0.76 \\
PI$_{0.5}$-LIBERO-10 & 0.89 & 0.50 & 0.79 & 0.88 & 0.71 & 0.88 & 0.89 & \textbf{0.90} & 0.89 & 0.87 \\
PI$_{0.5}$-ALOHA-real-drawer & 0.76 & 0.50 & 0.69 & 0.82 & 0.70 & 0.78 & 0.84 & \textbf{0.87} & 0.73 & 0.78 \\
\textit{ID average} & 0.83 & 0.58 & 0.77 & 0.84 & 0.70 & 0.78 & 0.85 & \textbf{0.87} & 0.80 & 0.80 \\
\addlinespace
PI$_{0.5}$-LIBERO $\rightarrow$ PI$_{0.5}$-ALOHA-real-block & 0.70 & 0.50 & 0.50 & 0.50 & 0.59 & 0.50 & 0.62 & 0.71 & 0.50 & \textbf{0.79} \\
PI$_0$-LIBERO $\rightarrow$ PI$_{0.5}$-LIBERO & \textbf{0.90} & 0.50 & 0.50 & 0.50 & 0.50 & 0.50 & 0.65 & 0.87 & 0.87 & 0.85 \\
PI$_{0.5}$-LIBERO $\rightarrow$ ACT-ALOHA-real-cup & \textbf{0.82} & 0.50 & 0.50 & 0.50 & 0.50 & 0.50 & 0.50 & 0.50 & 0.50 & 0.50 \\
\textit{OOD average} & \textbf{0.81} & 0.50 & 0.50 & 0.50 & 0.53 & 0.50 & 0.59 & 0.69 & 0.62 & 0.71 \\
\bottomrule
\end{tabular}
}
\end{table}

\vspace{0.5em}
\noindent\textbf{Comparison against action-based scores.}
Table~\ref{tab:action_score_results} and
Figure~\ref{fig:appendix_action_scores} compare Tri-Info against FIPER-ACE and
the four action-continuity scores. In domain, the alternatives are competitive
and two of them are stronger than Tri-Info: on raw embeddings, kNN distance
averages $0.84$ mBA against $0.83$ for Tri-Info, and on the matched latent kNN
and variance reach $0.87$ and $0.85$. FIPER-ACE is the weakest at $0.58$ and is
unchanged by the projection, since it reads sampled action chunks rather than
embeddings. In-domain accuracy therefore does not favour the
information-theoretic form: action dispersion in a well-scaled latent already
suffices.

The methods separate under shift, and the projection largely determines how far.
On raw embeddings, variance, kNN, and gradient fall to chance on every transfer,
and cosine dissimilarity rises above chance only cross-environment ($0.59$),
giving OOD averages of $0.50$ to $0.53$. These scores are expressed in the units
of one policy's embedding space, so a threshold fitted on the source does not
carry over. The matched latent recovers part of their accuracy, raising the OOD
average to $0.71$ for gradient and $0.69$ for kNN. The recovery comes entirely
from the two milder transfers. Cross-environment, gradient reaches $0.79$ and
kNN $0.71$, both above Tri-Info's $0.70$. Cross-model, kNN and cosine
dissimilarity reach $0.87$ and gradient $0.85$, all below Tri-Info's $0.90$. On
the transfer that changes model, embodiment, action dimensionality, and
appearance at once, all four return to $0.50$ in both representations, while
Tri-Info holds $0.82$. Tri-Info is therefore not the strongest on every
transfer, but it is the only method in Table~\ref{tab:action_score_results} that
stays above chance on all three, and its OOD average ($0.81$) exceeds the best
alternative ($0.71$) by $0.10$.

We draw two conclusions. The VAE projection accounts for a substantial share of
whatever transfer an action statistic achieves, which is why we report both
variants rather than attributing transfer to the metric alone. 
The projection is
nevertheless not sufficient: Once the latent is matched, entropy and mutual information remain comparable across domains because they describe the distribution of actions (how probability mass is spread and how strongly actions depend on their history and states).

\subsection{Hyperparameter Search}
\label{app:hyperparameters}

Every method is tuned by an independent grid search, so that no comparison is
confounded by unequal tuning effort. Table~\ref{tab:hp} lists the grids
and the selected configurations. Each search runs on
the in-domain train/validation split of its own domain, and the selected
configuration is applied unchanged to the transfer settings: the cross-model
transfer inherits the PI$_0$-LIBERO configuration, the two ALOHA transfers the
PI$_{0.5}$-LIBERO one. Model hyperparameters are ranked by AUC, and $\alpha$,
which does not enter AUC, by mean balanced accuracy among the AUC-optimal
configurations.

\subsection{Additional Results and Sensitivity Analyses}
\label{app:additional-results}

This section reports confidence intervals on the OOD results and the sensitivity of Tri-Info to detector capacity and to the flow-matching integration step.

\vspace{0.5em}
\noindent\textbf{Confidence intervals on OOD.}
We add trajectory-level 95\% CIs to the main OOD and real-world curves. 
We resample the $n$ test trajectories with replacement, stratified by outcome
($B=2000$). 
Each cell reports balanced accuracy with its 95\% CI in bracket:

\par{\centering
\resizebox{0.99\textwidth}{!}{%
\begin{tabular}{lcccccc}
\toprule
Panel (OOD) & $n$ & 20\% & 40\% & 60\% & 80\% & 100\% \\
\midrule
PI$_0$-LIBERO-10 $\to$ PI$_{0.5}$-LIBERO-10
 & 180 & 0.91 [0.86,0.95] & 0.92 [0.89,0.95] & 0.92 [0.88,0.94] & 0.92 [0.88,0.94] & 0.91 [0.88,0.94] \\
PI$_{0.5}$-LIBERO-10 $\to$ PI$_{0.5}$-ALOHA-real-block
 & 60 & 0.61 [0.55,0.69] & 0.65 [0.57,0.72] & 0.71 [0.63,0.79] & 0.76 [0.68,0.85] & 0.83 [0.74,0.91] \\
PI$_{0.5}$-LIBERO-10 $\to$ ACT-ALOHA-real-cup
 & 120 & 0.80 [0.74,0.86] & 0.95 [0.91,0.98] & 0.90 [0.85,0.95] & 0.82 [0.75,0.88] & 0.78 [0.72,0.85] \\
\bottomrule
\end{tabular}
}
\par}

\vspace{0.5em}
\noindent\textbf{Detector capacity.}
To verify that our results do not hinge on detector capacity, we run a post-hoc
sweep over the GRU hidden size $N$ and the dropout rate on the two OOD transfers
(Table~\ref{tab:gru-capacity}). AUC varies within $0.818$--$0.839$ on the
cross-model transfer and $0.852$--$0.910$ on the sim-to-real transfer, with no
monotone trend in either hyperparameter. 

\begin{table}[h]
\centering\footnotesize\setlength{\tabcolsep}{5pt}
\caption{Detector capacity sweep. Trajectory-level AUC (mean$_{\pm\text{std}}$ over three seeds) under distribution shift, for joint settings of GRU hidden size $N$ and dropout. Detection is insensitive to both. Best per row in bold.}
\label{tab:gru-capacity}
\begin{tabular}{@{}lccccc@{}}
\toprule
AUC (OOD) $\backslash$ $(N,\text{dropout})$ & (32, 0.1) & (64, 0.2) & (128, 0.2) & (256, 0.1) & (256, 0.3) \\
\midrule
PI$_0$-LIBERO-10 $\to$ PI$_{0.5}$-LIBERO-10 & 0.818$_{\pm.006}$ & 0.824$_{\pm.002}$ & \textbf{0.839}$_{\pm.005}$ & 0.825$_{\pm.007}$ & 0.830$_{\pm.008}$ \\
PI$_{0.5}$-LIBERO-10 $\to$ ACT-ALOHA-real & 0.891$_{\pm.014}$ & 0.852$_{\pm.048}$ & \textbf{0.910}$_{\pm.020}$ & 0.888$_{\pm.029}$ & 0.862$_{\pm.016}$ \\
\bottomrule
\end{tabular}
\end{table}

\vspace{0.5em}
\noindent\textbf{Flow-matching integration step.}
Flow-matching VLAs have no discrete action decoder, so we read the action embedding $\mathbf{a}_t$ from the velocity field at the last integration step (Section~\ref{sec:method}). To check that this choice is not critical, we rerun the Tri-Info pipeline on PI$_{0.5}$-LIBERO, changing only the step $i\in\{0,\dots,9\}$ from which $\mathbf{a}_t$ is read (Table~\ref{tab:flow-step}).
mBA stays within $0.862$--$0.893$ (mean $0.881$, std $0.011$) with no trend across steps, and the default last step is within $0.004$ of the best. Detection is therefore insensitive to this choice; we keep the last step, which is closest to the executed action.

\begin{table}[h]
\centering
\small
\setlength{\tabcolsep}{5pt}
\caption{mBA versus the integration step used for the action embedding (PI$_{0.5}$-LIBERO). Default in bold; best underlined.}
\label{tab:flow-step}
\begin{tabular}{@{}lcccccccccc@{}}
\toprule
Integration step $i$ & 0 & 1 & 2 & 3 & 4 & 5 & 6 & 7 & 8 & 9 (default) \\
\midrule
mBA & 0.886 & 0.863 & 0.889 & 0.862 & \underline{0.893} & 0.880 & 0.874 & 0.886 & 0.883 & \textbf{0.889} \\
\bottomrule
\end{tabular}
\end{table}

\paragraph{Running time.}
We measure the detector's per-inference cost end-to-end, averaged over
$10{\times}30$ trajectories (Table~\ref{tab:window_runtime}). The cost rises
mildly with $W$ and reaches at most 5.30\,ms on H800 (5.06\,ms on RTX~4090)
even at $W{=}50$, a small fraction of a typical VLA inference step, so the
detector adds negligible overhead and does not perturb the control loop.
\begin{table}[t]
\caption{Detector latency grows only mildly with window size $W$. Per-inference latency (ms), averaged over $10{\times}30$ trajectories.}
\label{tab:window_runtime}
\centering
\small
\setlength{\tabcolsep}{6pt}
\begin{tabular}{@{}llccccc@{}}
\toprule
 & & \multicolumn{5}{c}{Window size $W$} \\
\cmidrule(l){3-7}
GPU & Model & 10 & 20 & 30 & 40 & 50 \\
\midrule
\multirow{2}{*}{H800}     & PI$_0$     & 2.64 & 3.14 & 3.65 & 4.54 & 5.22 \\
                          & PI$_{0.5}$ & 2.69 & 3.21 & 3.72 & 4.61 & 5.30 \\
\midrule
\multirow{2}{*}{RTX 4090} & PI$_0$     & 4.08 & 4.27 & 4.47 & 4.79 & 5.06 \\
                          & PI$_{0.5}$ & 4.02 & 4.22 & 4.42 & 4.74 & 5.01 \\
\bottomrule
\end{tabular}
\end{table}

\noindent\textbf{Threshold calibration: $\alpha$ trades timeliness for accuracy.}
The significance level $\alpha$ sets the detector's tolerance to transient anomalies (Table~\ref{tab:threshold-TNR}). 
\begin{wraptable}{r}{0.42\textwidth}
\caption{$\alpha$ controls tolerance for transient anomalies. TNR: fraction of successful trajectories not falsely flagged.}
\label{tab:threshold-TNR}
\centering
\small
\setlength{\tabcolsep}{5pt}
\begin{tabular}{@{}l cccc@{}}
\toprule
$\alpha$ & TPR & TNR & Bal.~Acc & Det.\% \\
\midrule
\rowcolor{bestrow}
0.010 & .958 & \best{.906} & \best{.932} & 16.0 \\
0.050 & .995 & .736 & .865 & 9.6 \\
0.100 & .995 & .584 & .789 & 5.5 \\
\bottomrule
\end{tabular}
\vspace{-1.5em}
\end{wraptable}Because the threshold is the $(1{-}\alpha)$ quantile of successful-trajectory scores, and successful trajectories that recover from mid-execution perturbations still carry the label $f^{\xi}{=}0$, a smaller $\alpha$ raises the threshold and lets more transient excursions pass without an alarm. Lowering $\alpha$ from 0.10 to 0.01 thus raises the true-negative rate from 0.584 to 0.906, while the true-positive rate stays above 0.95. The cost is timeliness: the first alarm moves from 5.5\% to 16.0\% of the trajectory. Even at the strictest setting, the detector reaches 0.932 balanced accuracy after only 16\% of execution, and practitioners can tune $\alpha$ to their risk profile.

\begin{table*}[t]
\centering
\caption{\textbf{Hyperparameter search.} (a) The grid swept for each method.
(b) The configuration selected in each of the three domains on which the search
is run. $h$: hidden size; $E$: training epochs; $W$: window length; $\alpha$: significance level; $\mathcal{A}=\{0.001,0.05,0.10,0.15\}$. Methods in
the upper block read model embeddings and are swept separately on the raw
embeddings (raw) and on the VAE-matched latent (VAE); methods in the lower block
read executed or sampled actions and are swept once, so their selected
configurations hold under both representations. FIPER-ACE fixes $W=30$, stride
$1$, and the translational dimension set.}
\label{tab:hp}
\small
\renewcommand{\arraystretch}{1.15}

\setlength{\tabcolsep}{6pt}
\begin{tabular}{@{}ll@{}}
\multicolumn{2}{@{}l}{\textbf{(a) Search grids}}\\[2pt]
\toprule
\textbf{Method} & \textbf{Grid} \\
\midrule
\rowcolor{black!4} \multicolumn{2}{@{}l}{\textit{Representation-dependent (swept once per representation)}}\\
SAFE & backbone $\times E \times \alpha$: $\{$mlp, lstm$\} \times \{50,100\} \times \mathcal{A}$ \\
Fail-Detect & $h \times \alpha$: $\{32,64,256\} \times (\mathcal{A} \cup \{0.20\})$ \\
FIPER & $\alpha \times$ combine $\times\, w_{\text{act}}$: $(\mathcal{A} \cup \{0.25\}) \times \{$and, or$\} \times \{10,15,20,45\}$ \\
GRU-Variance / kNN / CosDis / Gradient & $h \times E \times W \times \alpha$: $\{32,64,256\} \times \{50,100\} \times \{20,30\} \times \mathcal{A}$ \\
\midrule
\rowcolor{black!4} \multicolumn{2}{@{}l}{\textit{Representation-independent (swept once)}}\\
STAC & score mode $\times\, \alpha$: $\{$cumulative, max$\} \times \mathcal{A}$ \\
GRU-STAC & $h \times E \times W \times \alpha$: $\{32,64,256\} \times \{50,100\} \times \{20,30\} \times \mathcal{A}$ \\
FIPER-ACE & aggregation $\times\, h \times E$: 4 settings $\times \{32,64,256\} \times \{50,100\}$ \\
Tri-Info & $h \times E \times \alpha$: $\{32,64,256\} \times \{50,100\} \times \mathcal{A}$ \\
\bottomrule
\end{tabular}

\vspace{1.1em}

\setlength{\tabcolsep}{4pt}
\resizebox{\textwidth}{!}{%
\begin{tabular}{@{}lllccc@{}}
\multicolumn{6}{@{}l}{\textbf{(b) Selected configurations}}\\[2pt]
\toprule
& & & \multicolumn{3}{c}{\textbf{Domain}} \\
\cmidrule(l){4-6}
\textbf{Method} & \textbf{Format} & \textbf{Repr.} & PI$_0$-LIBERO-10 & PI$_{0.5}$-LIBERO-10 & PI$_{0.5}$-ALOHA-real-drawer \\
\midrule
 & & raw & lstm, 100, .05 & mlp, 50, .15 & mlp, 50, .10 \\
\rowcolor{black!4} \multirow{-2}{*}{SAFE} & \multirow{-2}{*}{(backbone, $E$, $\alpha$)} & VAE & lstm, 100, .10 & mlp, 100, .15 & lstm, 100, .05 \\
\addlinespace[1pt]
 & & raw & 256, .15 & 256, .20 & 256, .15 \\
\rowcolor{black!4} \multirow{-2}{*}{Fail-Detect} & \multirow{-2}{*}{$(h, \alpha)$} & VAE & 256, .15 & 256, .15 & 256, .10 \\
\addlinespace[1pt]
 & & raw & .05, or, 15 & .05, or, 15 & .05, or, 45 \\
\rowcolor{black!4} \multirow{-2}{*}{FIPER} & \multirow{-2}{*}{$(\alpha, \text{combine}, w_{\text{act}})$} & VAE & .25, or, 45 & .05, or, 10 & .10, or, 10 \\
\addlinespace[1pt]
 & & raw & 256, 100, 20, .05 & 256, 100, 20, .05 & 256, 50, 20, .15 \\
\rowcolor{black!4} \multirow{-2}{*}{GRU-Variance} & \multirow{-2}{*}{$(h, E, W, \alpha)$} & VAE & 256, 100, 20, .05 & 32, 50, 20, .15 & 64, 50, 20, .05 \\
\addlinespace[1pt]
 & & raw & 256, 100, 20, .05 & 256, 100, 30, .05 & 256, 50, 30, .15 \\
\rowcolor{black!4} \multirow{-2}{*}{GRU-kNN} & \multirow{-2}{*}{$(h, E, W, \alpha)$} & VAE & 256, 100, 20, .05 & 64, 100, 20, .05 & 64, 50, 30, .15 \\
\addlinespace[1pt]
 & & raw & 64, 100, 30, .05 & 256, 100, 20, .05 & 64, 100, 30, .05 \\
\rowcolor{black!4} \multirow{-2}{*}{GRU-CosDis} & \multirow{-2}{*}{$(h, E, W, \alpha)$} & VAE & 256, 50, 30, .15 & 64, 50, 20, .05 & 256, 100, 30, .05 \\
\addlinespace[1pt]
 & & raw & 32, 50, 20, .15 & 64, 50, 20, .05 & 256, 50, 20, .15 \\
\rowcolor{black!4} \multirow{-2}{*}{GRU-Gradient} & \multirow{-2}{*}{$(h, E, W, \alpha)$} & VAE & 64, 100, 30, .05 & 32, 50, 20, .15 & 256, 100, 20, .05 \\
\midrule
Tri-Info & $(h, E, \alpha)$ & both & 32, 100, .10 & 256, 100, .10 & 32, 50, .05 \\
STAC & (mode, $\alpha$) & both & cumulative, .10 & max, .10 & cumulative, .10 \\
GRU-STAC & $(h, E, W, \alpha)$ & both & 32, 100, 20, .05 & 256, 100, 20, .15 & 32, 100, 20, .05 \\
FIPER-ACE & (aggregation, $h$, $E$) & both & window-30, 256, 50 & raw, 256, 100 & raw, 32, 50 \\
\bottomrule
\end{tabular}%
}
\end{table*}

\section{Limitations.} 
Firstly, the scope of this paper is mainly alarms, rather than recovery or improvement of the success rate.
We do try using the detector to select among sampled actions at inference time, which does
raise the success rate (Appendix~\ref{app:more-explorations}), though closing the loop deserves a
more systematic study than we give it here.
Secondly, this work is also tested only on VLAs; we additionally try a non-VLA behavioral-cloning policy
(Appendix~\ref{app:more-explorations}), which suggests the method may apply to closed-loop
policies in general, but confirming this requires a wider range of architectures.
 Finally, our approach relies on a VAE to compress high-dimensional embeddings into a low-dimensional latent space where entropy and mutual information can be estimated. 
Therefore, our method requires training a VAE to compress the embeddings.  Future work could remove this compression step by adopting information estimators that scale to high-dimensional spaces.

\section{Validating the Three-Signal Reduction}
\label{app:multi-feature}

\subsection{Exhaustive Subset Search}
\label{app:subset-search}

We cross-check the metric reduction with an
exhaustive search whose purpose is to confirm \emph{which} signals to retain.
We use logistic regression rather than the GRU: a memoryless classifier exposes
the intrinsic discriminative content of each metric, so the resulting ranking is
attributable to the metrics themselves rather than to the temporal model.
Scoring a subset instantaneously is also cheap enough to cover the whole
lattice, so we evaluate all $2^8-1=255$ non-empty subsets, fitting one logistic
regression per subset on per-timestep features pooled over all
environment-model combinations excluding OOD domain in Figure. \ref{fig:baseline_accuracy_results}. Table~\ref{tab:feature_results} reports the four
highest-AUC subsets at each cardinality.

The search independently recovers $\{\mathbb{H}(\mathbf{A}_t),\, \mathbb{I}(\mathbf{A}_t; \mathbf{A}_{t+1}),\, \mathbb{I}(\mathbf{S}_t, \mathbf{S}_{t+1}; \mathbf{A}_t)\}$ as the top-AUC subset of size~3 ($0.900$), though the AUC is very flat across ID performance. Performance is bounded at this point rather than improved by adding signals: size 4 and size 5 match $0.900$ without surpassing it, and size $\geq 6$ declines steadily to $0.876$ at size 8, as the remaining metrics are correlated with those already selected
(Figure~\ref{fig:mi_correlation}) and contribute redundancy rather than
complementary discrimination. 
The optimal subsets at every size $\geq 2$
retain the action-centric pair $\mathbb{I}(\mathbf{A}_t; \mathbf{A}_{t+1})$ and $\mathbb{H}(\mathbf{A}_t)$, with
$\mathbb{I}(\mathbf{S}_t, \mathbf{S}_{t+1}; \mathbf{A}_t)$ joining at size 3, consistent with our preference
for action-centric signals: successful execution maintains consistent and
diverse actions that remain coupled to state transitions.

\begin{table*}[t]
\centering
\caption{Detection performance AUC across different feature combinations. Best results per subset size are \textbf{bolded}.}
\label{tab:feature_results}
\resizebox{\textwidth}{!}{%
\begin{tabular}{@{}cl c@{}}
\toprule
Size & Features & AUC \\
\midrule
\multirow{4}{*}{2} 
& $\mathbb{H}(\mathbf{A}_t)$, $\mathbb{I}(\mathbf{A}_t; \mathbf{A}_{t+1})$ & \textbf{0.898} \\
& $\mathbb{I}(\mathbf{S}_t; \mathbf{S}_{t+1})$, $\mathbb{I}(\mathbf{A}_t; \mathbf{A}_{t+1})$ & 0.896 \\
& $\mathbb{I}(\mathbf{S}_{t+1}; \mathbf{A}_t)$, $\mathbb{I}(\mathbf{A}_t; \mathbf{A}_{t+1})$ & 0.895 \\
& $\mathbb{I}(\mathbf{S}_t, \mathbf{S}_{t+1}; \mathbf{A}_t)$, $\mathbb{I}(\mathbf{A}_t; \mathbf{A}_{t+1})$ & 0.895 \\
\midrule
\multirow{4}{*}{3} 
& $\mathbb{H}(\mathbf{A}_t)$, $\mathbb{I}(\mathbf{S}_t, \mathbf{S}_{t+1}; \mathbf{A}_t)$, $\mathbb{I}(\mathbf{A}_t; \mathbf{A}_{t+1})$ & \textbf{0.900} \\
& $\mathbb{H}(\mathbf{A}_t)$, $\mathbb{I}(\mathbf{A}_t, \mathbf{S}_t; \mathbf{S}_{t+1})$, $\mathbb{I}(\mathbf{A}_t; \mathbf{A}_{t+1})$ & 0.899 \\
& $\mathbb{H}(\mathbf{A}_t)$, $\mathbb{I}(\mathbf{S}_t; \mathbf{S}_{t+1})$, $\mathbb{I}(\mathbf{A}_t; \mathbf{A}_{t+1})$ & 0.899 \\
& $\mathbb{H}(\mathbf{A}_t)$, $\mathbb{I}(\mathbf{S}_{t+1}; \mathbf{A}_t)$, $\mathbb{I}(\mathbf{A}_t; \mathbf{A}_{t+1})$ & 0.898 \\
\midrule
\multirow{4}{*}{4} 
& $\mathbb{H}(\mathbf{A}_t)$, $\mathbb{I}(\mathbf{A}_t, \mathbf{S}_t; \mathbf{S}_{t+1})$, $\mathbb{I}(\mathbf{S}_t, \mathbf{S}_{t+1}; \mathbf{A}_t)$, $\mathbb{I}(\mathbf{A}_t; \mathbf{A}_{t+1})$ & \textbf{0.900} \\
& $\mathbb{H}(\mathbf{A}_t)$, $\mathbb{I}(\mathbf{S}_t, \mathbf{S}_{t+1}; \mathbf{A}_t)$, $\mathbb{I}(\mathbf{S}_t; \mathbf{S}_{t+1})$, $\mathbb{I}(\mathbf{A}_t; \mathbf{A}_{t+1})$ & 0.899 \\
& $\mathbb{H}(\mathbf{A}_t)$, $\mathbb{I}(\mathbf{A}_t, \mathbf{S}_t; \mathbf{S}_{t+1})$, $\mathbb{I}(\mathbf{S}_t; \mathbf{S}_{t+1})$, $\mathbb{I}(\mathbf{A}_t; \mathbf{A}_{t+1})$ & 0.899 \\
& $\mathbb{H}(\mathbf{A}_t)$, $\mathbb{I}(\mathbf{S}_{t+1}; \mathbf{A}_t)$, $\mathbb{I}(\mathbf{S}_t; \mathbf{S}_{t+1})$, $\mathbb{I}(\mathbf{A}_t; \mathbf{A}_{t+1})$ & 0.899 \\
\midrule
\multirow{4}{*}{5} 
& $\mathbb{H}(\mathbf{A}_t)$, $\mathbb{I}(\mathbf{A}_t, \mathbf{S}_t; \mathbf{S}_{t+1})$, $\mathbb{I}(\mathbf{S}_t, \mathbf{S}_{t+1}; \mathbf{A}_t)$, $\mathbb{I}(\mathbf{S}_t; \mathbf{S}_{t+1})$, $\mathbb{I}(\mathbf{A}_t; \mathbf{A}_{t+1})$ & \textbf{0.900} \\
& $\mathbb{H}(\mathbf{A}_t)$, $\mathbb{I}(\mathbf{S}_t; \mathbf{A}_t)$, $\mathbb{I}(\mathbf{S}_{t+1}; \mathbf{A}_t)$, $\mathbb{I}(\mathbf{S}_t; \mathbf{S}_{t+1})$, $\mathbb{I}(\mathbf{A}_t; \mathbf{A}_{t+1})$ & 0.899 \\
& $\mathbb{H}(\mathbf{A}_t)$, $\mathbb{I}(\mathbf{S}_t; \mathbf{A}_t)$, $\mathbb{I}(\mathbf{A}_t, \mathbf{S}_t; \mathbf{S}_{t+1})$, $\mathbb{I}(\mathbf{S}_t; \mathbf{S}_{t+1})$, $\mathbb{I}(\mathbf{A}_t; \mathbf{A}_{t+1})$ & 0.899 \\
& $\mathbb{H}(\mathbf{A}_t)$, $\mathbb{I}(\mathbf{S}_{t+1}; \mathbf{A}_t)$, $\mathbb{I}(\mathbf{A}_t, \mathbf{S}_t; \mathbf{S}_{t+1})$, $\mathbb{I}(\mathbf{S}_t; \mathbf{S}_{t+1})$, $\mathbb{I}(\mathbf{A}_t; \mathbf{A}_{t+1})$ & 0.894 \\
\midrule
\multirow{4}{*}{6} 
& $\mathbb{H}(\mathbf{A}_t)$, $\mathbb{I}(\mathbf{S}_t; \mathbf{A}_t)$, $\mathbb{I}(\mathbf{S}_{t+1}; \mathbf{A}_t)$, $\mathbb{I}(\mathbf{A}_t, \mathbf{S}_t; \mathbf{S}_{t+1})$, $\mathbb{I}(\mathbf{S}_t; \mathbf{S}_{t+1})$, $\mathbb{I}(\mathbf{A}_t; \mathbf{A}_{t+1})$ & \textbf{0.895} \\
& $\mathbb{H}(\mathbf{A}_t)$, $\mathbb{I}(\mathbf{S}_t; \mathbf{A}_t)$, $\mathbb{I}(\mathbf{A}_t, \mathbf{S}_t; \mathbf{S}_{t+1})$, $\mathbb{I}(\mathbf{S}_t, \mathbf{S}_{t+1}; \mathbf{A}_t)$, $\mathbb{I}(\mathbf{S}_t; \mathbf{S}_{t+1})$, $\mathbb{I}(\mathbf{A}_t; \mathbf{A}_{t+1})$ & 0.892 \\
& $\mathbb{H}(\mathbf{A}_t)$, $\mathbb{I}(\mathbf{S}_{t+1}; \mathbf{A}_t)$, $\mathbb{I}(\mathbf{A}_t, \mathbf{S}_t; \mathbf{S}_{t+1})$, $\mathbb{I}(\mathbf{S}_t, \mathbf{S}_{t+1}; \mathbf{A}_t)$, $\mathbb{I}(\mathbf{S}_t; \mathbf{S}_{t+1})$, $\mathbb{I}(\mathbf{A}_t; \mathbf{A}_{t+1})$ & 0.889 \\
& $\mathbb{H}(\mathbf{A}_t)$, $\mathbb{I}(\mathbf{S}_t; \mathbf{A}_t)$, $\mathbb{I}(\mathbf{S}_{t+1}; \mathbf{A}_t)$, $\mathbb{I}(\mathbf{S}_t, \mathbf{S}_{t+1}; \mathbf{A}_t)$, $\mathbb{I}(\mathbf{S}_t; \mathbf{S}_{t+1})$, $\mathbb{I}(\mathbf{A}_t; \mathbf{A}_{t+1})$ & 0.888 \\
\midrule
\multirow{4}{*}{7} 
& $\mathbb{H}(\mathbf{A}_t)$, $\mathbb{I}(\mathbf{S}_t; \mathbf{A}_t)$, $\mathbb{I}(\mathbf{S}_{t+1}; \mathbf{A}_t)$, $\mathbb{I}(\mathbf{A}_t, \mathbf{S}_t; \mathbf{S}_{t+1})$, $\mathbb{I}(\mathbf{S}_t, \mathbf{S}_{t+1}; \mathbf{A}_t)$, $\mathbb{I}(\mathbf{S}_t; \mathbf{S}_{t+1})$, $\mathbb{I}(\mathbf{A}_t; \mathbf{A}_{t+1})$ & \textbf{0.888} \\
& $\mathbb{H}(\mathbf{S}_t)$, $\mathbb{H}(\mathbf{A}_t)$, $\mathbb{I}(\mathbf{S}_t; \mathbf{A}_t)$, $\mathbb{I}(\mathbf{A}_t, \mathbf{S}_t; \mathbf{S}_{t+1})$, $\mathbb{I}(\mathbf{S}_t, \mathbf{S}_{t+1}; \mathbf{A}_t)$, $\mathbb{I}(\mathbf{S}_t; \mathbf{S}_{t+1})$, $\mathbb{I}(\mathbf{A}_t; \mathbf{A}_{t+1})$ & 0.879 \\
& $\mathbb{H}(\mathbf{S}_t)$, $\mathbb{I}(\mathbf{S}_t; \mathbf{A}_t)$, $\mathbb{I}(\mathbf{S}_{t+1}; \mathbf{A}_t)$, $\mathbb{I}(\mathbf{A}_t, \mathbf{S}_t; \mathbf{S}_{t+1})$, $\mathbb{I}(\mathbf{S}_t, \mathbf{S}_{t+1}; \mathbf{A}_t)$, $\mathbb{I}(\mathbf{S}_t; \mathbf{S}_{t+1})$, $\mathbb{I}(\mathbf{A}_t; \mathbf{A}_{t+1})$ & 0.877 \\
& $\mathbb{H}(\mathbf{S}_t)$, $\mathbb{H}(\mathbf{A}_t)$, $\mathbb{I}(\mathbf{S}_t; \mathbf{A}_t)$, $\mathbb{I}(\mathbf{S}_{t+1}; \mathbf{A}_t)$, $\mathbb{I}(\mathbf{S}_t, \mathbf{S}_{t+1}; \mathbf{A}_t)$, $\mathbb{I}(\mathbf{S}_t; \mathbf{S}_{t+1})$, $\mathbb{I}(\mathbf{A}_t; \mathbf{A}_{t+1})$ & 0.877 \\
\midrule
8 & $\mathbb{H}(\mathbf{S}_t)$, $\mathbb{H}(\mathbf{A}_t)$, $\mathbb{I}(\mathbf{S}_t; \mathbf{A}_t)$, $\mathbb{I}(\mathbf{S}_{t+1}; \mathbf{A}_t)$, $\mathbb{I}(\mathbf{A}_t, \mathbf{S}_t; \mathbf{S}_{t+1})$, $\mathbb{I}(\mathbf{S}_t, \mathbf{S}_{t+1}; \mathbf{A}_t)$, $\mathbb{I}(\mathbf{S}_t; \mathbf{S}_{t+1})$, $\mathbb{I}(\mathbf{A}_t; \mathbf{A}_{t+1})$ & 0.876 \\
\bottomrule
\end{tabular}%
}
\end{table*}

\begin{table}
\centering
\caption{Action and state variational autoencoder (VAE) encoder architectures. Both branch into parallel
mean and log-variance heads at a matched $d_z$-dim latent ($d_z = 7$); decoders mirror the
encoders with a sigmoid output. The VAE state encoder runs once per input modality
(images, proprioception); $n_{\mathrm{mod}}$ denotes the number of input
modalities, which varies across models, and $d_{\mathrm{int}}$ the per-modality
intermediate code width ($d_{\mathrm{int}} = 8$), so each modality head emits
$2 d_{\mathrm{int}}$ values for its own mean/log-variance pair.}
\label{tab:vae_arch}
\begin{tabular}{@{}ll@{}}
\toprule
Action VAE & State VAE (per modality) \\
\midrule
$\mathrm{Linear}(\to 500)$        & $\mathrm{Linear}(\to \min(512,\, \text{in}/4))$ \\
$\mathrm{LeakyReLU}$              & $\mathrm{LayerNorm} + \mathrm{LeakyReLU} + \mathrm{Dropout}$ \\
$\mathrm{Linear}(500 \to 500)$    & $\mathrm{Linear}(\to 2 d_{\mathrm{int}})$ \\
$\mathrm{LeakyReLU}$              & Concatenate $n_{\mathrm{mod}}$ modalities $\to d_{\mathrm{int}}\,n_{\mathrm{mod}}$ \\
                                  & $\mathrm{Linear}(d_{\mathrm{int}}\,n_{\mathrm{mod}} \to 64) + \mathrm{LayerNorm} + \mathrm{LeakyReLU}$ \\
Mean / log-var heads: $500 \to d_z$ & Mean / log-var heads: $64 \to d_z$ \\
\bottomrule
\end{tabular}
\end{table}

\section{MI Estimation}
\label{app:mi_estimation}

We estimate the three Tri-Info quantities $\mathbb{H}(\mathbf{A}_t)$,
$\mathbb{I}(\mathbf{A}_t;\mathbf{A}_{t+1})$, and $\mathbb{I}(\mathbf{S}_t,\mathbf{S}_{t+1};\mathbf{A}_t)$
non-parametrically over a sliding window of length $W$, using the
Kozachenko--Leonenko estimator~\citep{kozachenko1987sample} for the entropy and
the Kraskov--St\"ogbauer--Grassberger estimator~\citep{kraskov2004estimating} for
the mutual-information terms, both under the max-norm with $k=5$ neighbours and
$N=W-1$ samples per window.

\paragraph{VAE projection.}
Raw embeddings are too high-dimensional for $k$-NN estimation, where neighbour
distances concentrate and the estimator bias grows. Following
\citet{hejna2025robot}, who estimate state--action mutual information with the
same estimator family, we compress each embedding to a matched $d_z{=}7$ latent
with a VAE fitted per model--dataset domain (Table~\ref{tab:vae_arch};
Appendix~\ref{app:protocol}) and estimate on the posterior means. A VAE rather
than an arbitrary projection is what makes the distances usable: its isotropic
Gaussian prior gives the latent a well-conditioned geometry, a prerequisite for
$k$-NN statistics and the regime in which these estimators are best
characterized. Keeping $d_z$ small serves the estimator, as they argue, and
matching it across the two encoders keeps the joint dimensions of the three
metrics comparable; their ablation over a threefold range of latent sizes
reports little effect, and we did not repeat that sweep here.

\paragraph{Assumptions and what the estimates support.}
Our windowed use stresses the estimators more than the pooled use they were
validated under: \citet{hejna2025robot} draw batches of $1024$ samples from a
whole dataset, whereas a window affords $29$ consecutive -- hence autocorrelated
-- steps at the default $W=30$, and they take the $L_2$ norm within each latent
block and the max-norm between blocks, where we use the max-norm in every
subspace. Local uniformity inside the $k$-NN ball is correspondingly loose,
across up to $21$ joint dimensions for
$\mathbb{I}(\mathbf{S}_t,\mathbf{S}_{t+1};\mathbf{A}_t)$, and the entropy estimator
diverges as neighbour distances vanish, so we floor them at $10^{-15}$: a
degenerate window yields a finite floor for $\mathbb{H}(\mathbf{A}_t)$ rather than a
divergence. Absolute values in nats are therefore not meaningful and we never
use them as such; as in \citet{hejna2025robot}, only relative comparisons carry
weight. We accordingly read each metric as a normalized time series
 whose shape rather than level drives the GRU, and
let the detector integrate the sequence of per-window estimates rather than
trust any one of them. This is also why $W$ is the one consequential
hyperparameter below -- it sets both the sample count and the span over which
the regime must be stationary -- while detection is insensitive to $k$, matching
the robustness \citet{hejna2025robot} report over $k\in\{2,\ldots,10\}$.

\begin{figure}[t]
    \centering
    \includegraphics[width=0.49\linewidth]{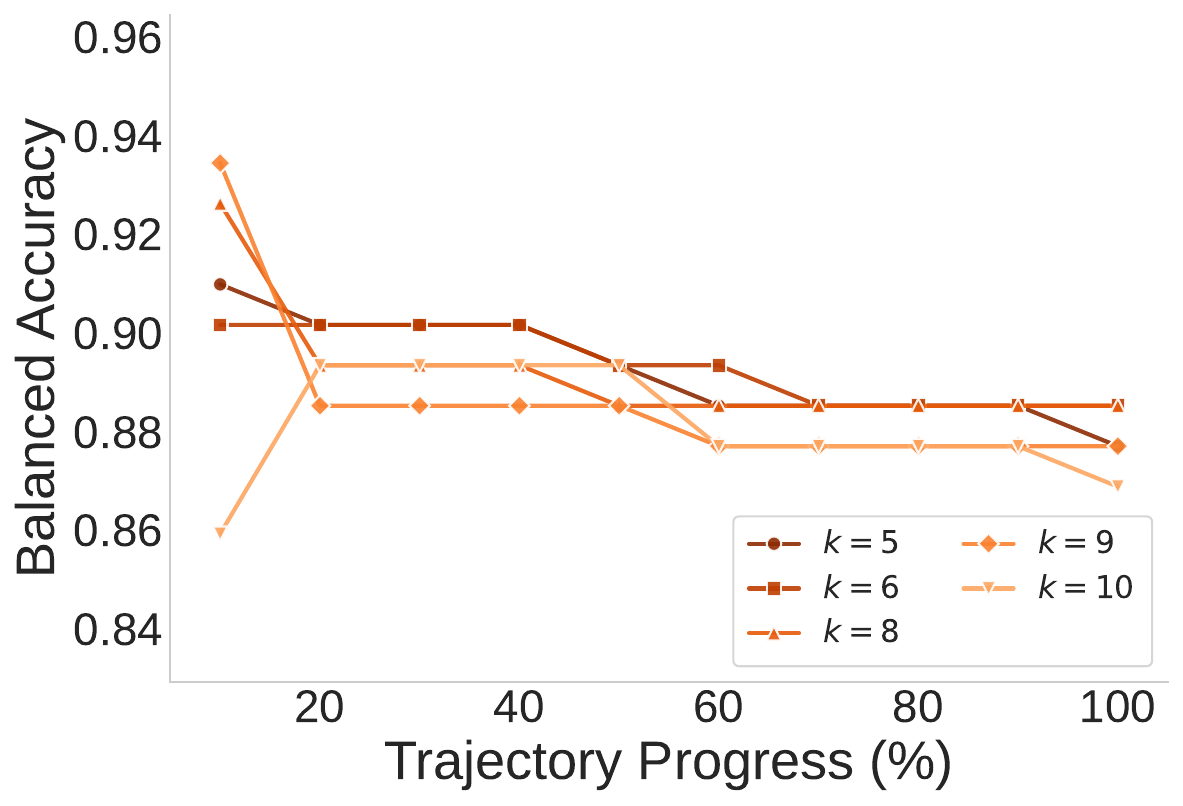}
    \hfill
    \includegraphics[width=0.49\linewidth]{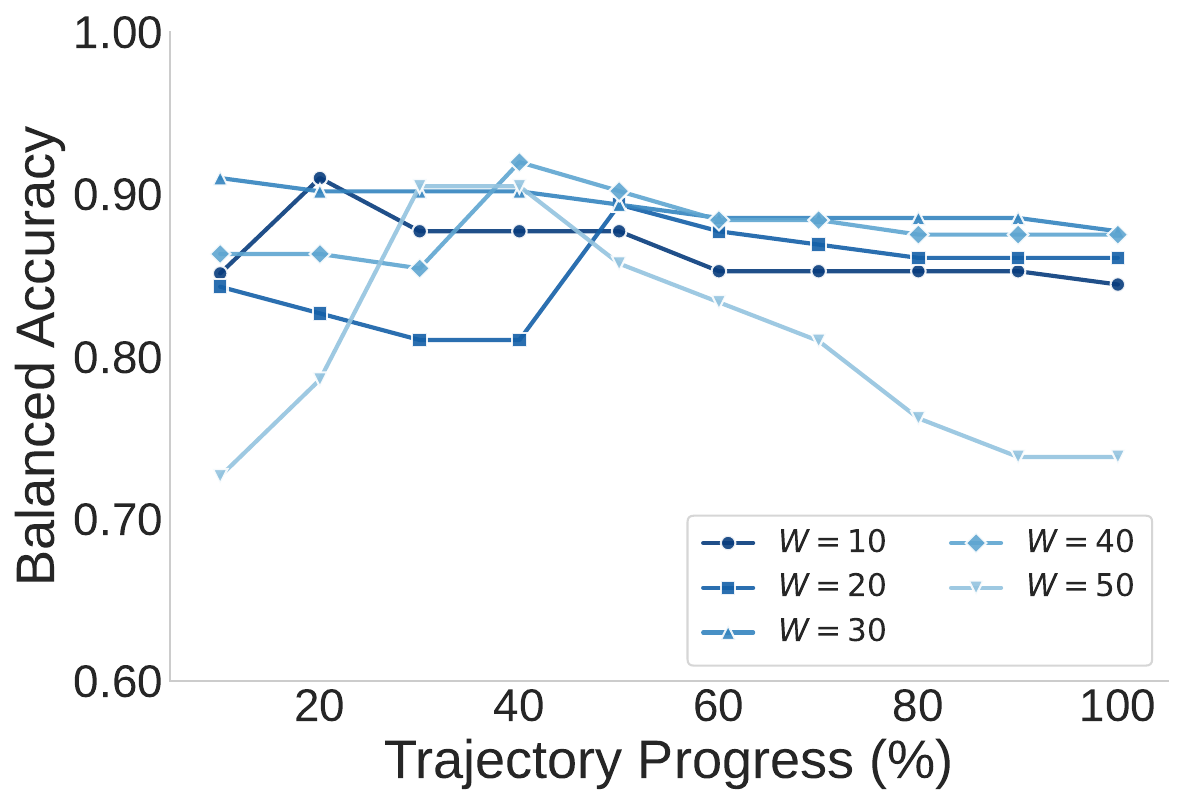}
    \caption{Balanced accuracy versus trajectory progress for the Tri-Info detector (PI$_{0.5}$-LIBERO) under the two estimator hyperparameter sweeps. \textbf{Left:} neighborhood size $k\in\{5,6,8,9,10\}$ at fixed $W=30$; the curves are tightly clustered, indicating that detection is insensitive to $k$. \textbf{Right:} window length $W\in\{10,20,30,40,50\}$ at fixed $k=5$; the wider spread, with $W=30$ highest, shows that $W$ is the more consequential hyperparameter.}
    \label{fig:wk_sweep}
\end{figure}

\paragraph{Sensitivity to Window Size $W$ and Neighborhood Size $k$}
\label{app:window_ablation}
We sweep the two estimator hyperparameters on the deployed detector (Tri-Info
with mean fusion on PI$_{0.5}$), varying $W$ and $k$ one at a time around their
defaults (Table~\ref{tab:wk_ablation}, Figure~\ref{fig:wk_sweep}). Detection is
largely insensitive to $k$, spanning $0.893$--$0.918$ AUC and $0.881$--$0.893$
BA over the sweep, with the default $k=5$ within $0.003$ AUC of the best value.
$W$ is the consequential choice: AUC stays within $0.875$--$0.915$ for
$W\leq 40$ and falls to $0.683$ at $W=50$, where BA also drops to $0.806$, since
long windows blur the failure transient the detector reads. The progress curves
echo this, spreading under the $W$ sweep and clustering under the $k$ sweep. We
therefore use $W=30$, the peak of both measures, and leave $k=5$ at its robust
default.

\begin{table}[t]
\centering
\caption{Sensitivity to the window size $W$ and the neighborhood size $k$ of the $k$-NN estimators. Each sweep varies one while holding the other at its default ($k=5$, $W=30$). Best per row in bold.}
\label{tab:wk_ablation}
\small
\setlength{\tabcolsep}{4pt}
\begin{minipage}[t]{0.45\linewidth}
\centering
\medskip
\begin{tabular}{lccccc}
\toprule
$k$ & 5 & 6 & 8 & 9 & 10 \\
\midrule
AUC & 0.915 & 0.893 & \textbf{0.918} & \textbf{0.918} & 0.893 \\
Avg. BA & \textbf{0.893} & \textbf{0.893} & 0.892 & 0.886 & 0.881 \\
\bottomrule
\end{tabular}
\end{minipage}\hfill
\begin{minipage}[t]{0.45\linewidth}
\centering
\medskip
\begin{tabular}{lccccc}
\toprule
$W$ & 10 & 20 & 30 & 40 & 50 \\
\midrule
AUC & 0.913 & 0.888 & \textbf{0.915} & 0.875 & 0.683 \\
Avg. BA & 0.865 & 0.851 & \textbf{0.893} & 0.879 & 0.806 \\
\bottomrule
\end{tabular}
\end{minipage}
\end{table}

\end{document}